\documentclass[11pt]{article}

\usepackage[preprint]{acl}

\usepackage{times}
\usepackage{latexsym}

\usepackage[T1]{fontenc}

\usepackage[utf8]{inputenc}

\usepackage{microtype}

\usepackage{inconsolata}
\usepackage{booktabs}
\usepackage{graphicx}
\usepackage{subcaption}
\usepackage{amsmath}
\usepackage{multirow,multicol}
\usepackage[inkscapeformat=png]{svg}

\usepackage[normalem]{ulem} 
\usepackage[table,usenames,dvipsnames]{xcolor}

\usepackage{tikz}
\newcommand*\circled[1]{\tikz[baseline=(char.base)]{
            \node[shape=circle,draw,inner sep=0.5pt] (char) {#1};}}
\def\circle#1{\circled{#1}}

\usepackage[ruled,vlined]{algorithm2e}

\usepackage[inline]{enumitem}
\usepackage[noabbrev,capitalize,nameinlink]{cleveref}
\usepackage{soul}
\usepackage{pbox}
\usepackage{longtable}
\usepackage{tablefootnote}
\usepackage{lscape}
\usepackage{ragged2e}
\usepackage{listings}       
\usepackage{todonotes}
\usepackage{supertabular}

\usepackage[normalem]{ulem} 
\usepackage{xcolor}

\usepackage{csvsimple}

\usepackage{amsmath} 

\AddToHook{cmd/appendix/before}{%
    \crefalias{section}{appendix}%
    \crefalias{subsection}{appendix}
}

\usepackage[english]{babel}
\addto\extrasenglish{
  
}

\usepackage[htt]{hyphenat}
\usepackage{makecell}

\title{Instability of LLM Pre-Pretraining: \ul{It Doesn't Always Help}.\\An Investigation on Multiple Languages}

\author{
 \hfill
 \textbf{Sofiia Riazhskykh}$^\dagger$ \hfill
 \textbf{Nam Luu}$^\dagger$ \hfill
 \textbf{Ond\v{r}ej Bojar} \hfill
 \hfill
\\
\\
 Charles University, Faculty of Mathematics and Physics
\\
 \texttt{sofiia.riazhskykh890@student.cuni.cz, \{luu,bojar\}@ufal.mff.cuni.cz}
 \\
 {\footnotesize $^\dagger$ Equal contribution}
}

\begin{document}
\maketitle
\begin{abstract}
Pretraining LLMs on artificial languages (``pre-pretraining'') is a technique that could reportedly increase token efficiency by 33\%, i.e., save up to 33\% of training tokens needed to reach a certain performance. We validate this prior result for English on a larger set of natural languages across four language families, using two different tokenizers and varying model sizes. We also relate the observed gains (or losses) in token efficiency to quantified linguistic properties of the languages, such as sentence length, morphological richness, and features of dependency syntactic trees (tree depth,  number of children, number of crossing dependencies). Our empirical results indicate that the reported gains
depend heavily on the experiment setup and the choice of random seed, although we can confirm the trend of stable gains with 128-Dyck pretraining of small models with the Llama tokenizer
for most of the examined languages.
On a general note, we argue that multiple training runs should be carried out at least for a subset of experiments to avoid the community adopting unstable approaches.
\end{abstract}

\section{Introduction}
\label{sec:1}

Language models require more training data as their size grows, which poses a challenge for low-resource languages and specific domains. Transfer learning, i.e., benefiting from the knowledge in a different dataset, is one of the established techniques for this situation. Previous work in the machine translation area \citep{kocmi-bojar-2018-trivial,lin-etal-2019-choosing} have shown that initial training on a related or even fully unrelated language pairs improves performance on the language pair of interest.
Later, gains were also observed when pretraining large language models (LLMs) on artificial languages (or ``pre-pretraining'') \citep{ri-tsuruoka-2022-pretraining,papadimitriou-jurafsky-2023-injecting,hu-etal-2025-circuits}. 
However, these recent studies on LLMs were solely focused on English.

To our knowledge, it remains unclear whether the method consistently improves token efficiency---i.e., reduces the amount of data needs to achieve a given performance---across other natural languages, model sizes, and architectures.

In this work, we go beyond the previous studies of LLM transfer learning from artificial languages in two main directions: we extend (1) the set of formal languages used, and (2) the set of natural languages of different families, seeking a relation between the properties of the artificial and natural languages affecting the effectiveness of the method.

\begin{figure}[t!]
    \centering
    \includegraphics[width=1\linewidth]{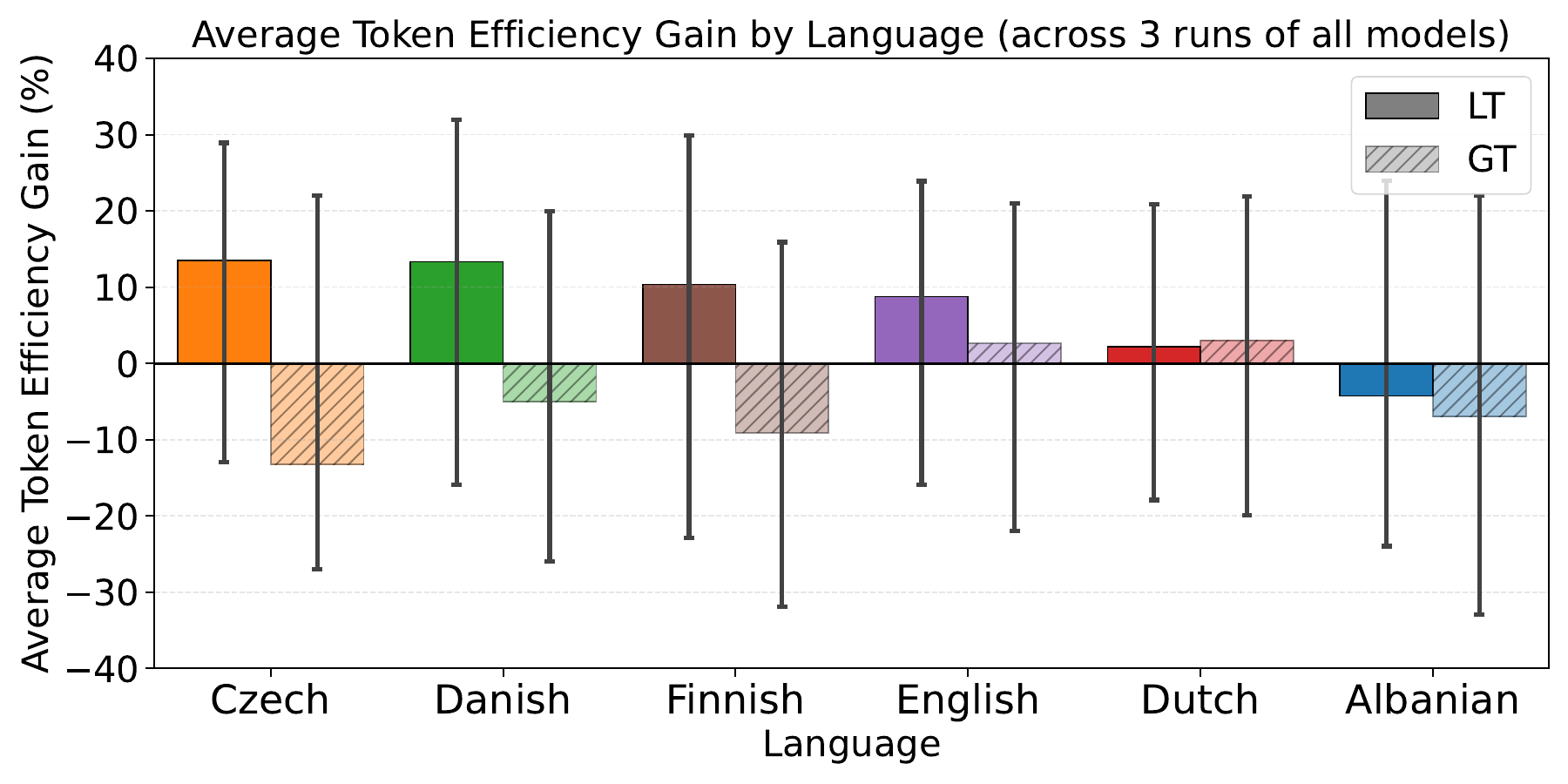}
    \vspace{-20pt}
    \caption{Average token efficiency gains of all models with two different tokenizers, where \texttt{LT} stands for Llama, and \texttt{GT} stands for Gemma tokenizer.}
    \label{fig:tokenizers_comparison}
    \vspace{-10pt}
\end{figure}

Experimental results suggest that the pretraining method may help across different natural languages, although the observed gains are highly dependent on the particular training run and are also sensitive to the setup details (\cref{fig:tokenizers_comparison}).


\section{Related Work}
\label{sec:related_work}

\citet{ri-tsuruoka-2022-pretraining} show that training LSTM- and Transformer-based language models on artificial languages can provide knowledge transfer to natural ones. They conclude \texttt{Nesting Dependency} outperforms \texttt{Flat Dependency} (see \cref{subsec:formal_languages}) in importance for language modeling. However, in contrast to the findings of \citet{ri-tsuruoka-2022-pretraining},
\citet{kumar-etal-2025-pretraining} observe that pretraining on \texttt{CROSS} outperforms pretraining on \texttt{NEST}.\footnote{\texttt{NEST} is equivalent to \texttt{Nesting Dependency}; \texttt{CROSS} is equivalent to \texttt{Flat Dependency}
}

Subsequent studies in \citet{papadimitriou-jurafsky-2023-injecting} and the ACL outstanding paper by \citet{hu-etal-2025-circuits} claim that applying the same method to a decoder-only LLM can improve its language acquisition, reducing perplexity for the Pythia 160M model, and also resulting in a 33\% token efficiency gain for the Pythia 1B variant (albeit with one run).



\citet{budnikov-yamshchikov-2025-transfer} suggest that some artificial languages used in pretraining lead to structurally richer token embeddings, but they fail to see any direct transfer of this to English.

\section{Dataset}
\label{sec:2}

\subsection{Formal Languages for Pretraining} \label{subsec:formal_languages}
Among formal languages, we consider $k$-Dyck (demonstrating \texttt{Nesting Dependency}) and $k$-Shuffle Dyck (\texttt{Flat Dependency}) languages for $k = 64$ and $k = 128$, which are defined as follows:

\textbf{$k$-Dyck} contains $k$ different types of structurally-nested dependencies, each of which is denoted by matching pair of opening and closing brackets, e.g., \texttt{{\color{blue} (} {\color{red} [} \{ \} {\color{red} ]} {\color{blue} )}} or \texttt{{\color{blue} (} {\color{red} [} {\color{red} ]} {\color{blue} )} \{ \}}. 

\textbf{$k$-Shuffle Dyck} allows breaking the well-nestedness (e.g., \texttt{{\color{blue} (} {\color{red} [} {\color{red} ]} \{ {\color{blue} )} \}}) and including cross-serial dependencies (e.g., \texttt{{\color{blue} (} {\color{red} [} \{ {\color{blue} )} {\color{red} ]} \}}).

We follow the approach of \citet{hu-etal-2025-circuits} and represent the bracket symbols with non-negative integers. 
In other words, if an opening symbol is denoted as the number $n$, its corresponding closing symbol is represented as $n + k$, where $k$ is the number of distinct dependencies, and $0 \leq n < k$. An example pair of opening and closing symbols for $k = 128$ is \texttt{(``81'', ``209'')}.
We refer readers to \citet{hu-etal-2025-circuits} for the exact implementation.

\subsection{Natural Languages}
\label{subsec:natural_languages}
Aside from English, we analyze the effects of pretraining for five other typologically diverse European languages from four language families, namely Albanian (Albanoid), Czech (Slavic), Danish and Dutch (Germanic), and Finnish (Uralic).

We construct the training dataset for each language as a monolingual, multi-domain dataset by mixing either FineWeb
(for English; \citealt{penedo2024the}) or FineWeb 2
(for the other five languages; \citealt{penedo2025fineweb2pipelinescale}) with OpenSubtitles
\cite{lison-tiedemann-2016-opensubtitles2016} and EUBookshop
\cite{TIEDEMANN12.463}, in proportions of 8:1:1, respectively.

Importantly, we attempt to isolate the effect of pretraining corpus size observed by \citet{kocmi-bojar-2018-trivial}: a larger pretraining corpus always delivers bigger gains. Across all experiments, our pretraining corpus consists of 16,000 Dyck sequences of 2,048 tokens (i.e., 500 training steps with the batch size of 32) and our natural language corpus used afterwards is 320,000 sequences\footnote{In line with the practice of training LLMs but different from machine translation area, our sequences span sentence boundaries.} of 2,048 tokens (i.e., 10,000 steps with the batch size of 32), subject to minor variations due to tokenization.

For evaluation, we use the first 2,000 rows of the validation split of the Multilingual Colossal Clean Crawled Corpus (mC4) dataset.\footnote{\href{https://huggingface.co/datasets/allenai/c4}{\texttt{allenai/c4}}} 


\section{Experiments and Results}
\label{sec:3}

\subsection{Experiment Setups}
\label{subsec:exp_setups}
We experiment with three model sizes, based on Llama 3 architecture \citep{grattafiori2024llama3herdmodels}. Each model is accompanied by either Llama 3 (\texttt{LT}) or Gemma 3 tokenizer (\texttt{GT}) \citep{gemmateam2025gemma3technicalreport}. This results in six different models, ranging from 253M to 884M. These models, along with the shorthand aliases used hereafter, are outlined in \cref{tab:models}.

\setlength{\tabcolsep}{3.5pt}
\begin{table}[h!]
\small
\centering
\begin{tabular}{|c|rrrc|}
\hline
\multirow{2}{*}{\textbf{Config}} & \multicolumn{3}{c|}{\textbf{\# of parameters}} & \multirow{2}{*}{\textbf{Name}}                                                                            \\ \cline{2-4} 
                                 & \multicolumn{1}{l|}{\textbf{Effective}} & \multicolumn{1}{l|}{\textbf{Embedding}} & \multicolumn{1}{l|}{\textbf{Total}} & \\ \hline \hline
\multirow{2}{*}{0}               & \multicolumn{1}{r|}{\multirow{2}{*}{154M}} & \multicolumn{1}{r|}{99M}                & \multicolumn{1}{r|}{253M} & \texttt{154M+LT}                              \\ \cline{3-5} 
                                 & \multicolumn{1}{r|}{}                      & \multicolumn{1}{r|}{201M}               & \multicolumn{1}{r|}{355M} & \texttt{154M+GT}                               \\ \hline
\multirow{2}{*}{1}               & \multicolumn{1}{r|}{\multirow{2}{*}{308M}} & \multicolumn{1}{r|}{131M}               & \multicolumn{1}{r|}{440M} & \texttt{308M+LT}                               \\ \cline{3-5} 
                                 & \multicolumn{1}{r|}{}                      & \multicolumn{1}{r|}{268M}               & \multicolumn{1}{r|}{577M} & \texttt{308M+GT}                               \\ \hline
\multirow{2}{*}{2}               & \multicolumn{1}{r|}{\multirow{2}{*}{481M}} & \multicolumn{1}{r|}{197M}               & \multicolumn{1}{r|}{678M} & \texttt{481M+LT}                               \\ \cline{3-5} 
                                 & \multicolumn{1}{r|}{}                      & \multicolumn{1}{r|}{403M}               & \multicolumn{1}{r|}{884M} & \texttt{481M+GT}                               \\ \hline
                                 
\end{tabular}
\caption{Overview of models. \texttt{LT} and \texttt{GT} stand for models with Llama and Gemma tokenizer, respectively.}
\label{tab:models}
\vspace{-10pt}
\end{table}

\begin{figure}[h!]
    \centering
    \includegraphics[width=\linewidth]{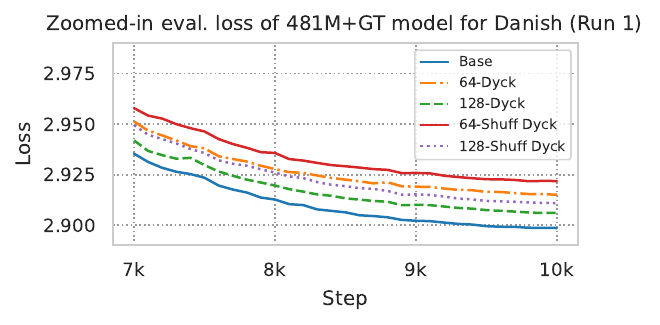}
    \vspace{-15pt}
    \caption{Evaluation loss of the \texttt{481M+GT} model for Danish, in the last 3,000 training steps. Lower is better.
    }
    \label{fig:eval_danish_481M_GT}
    \vspace{-10pt}
\end{figure}

With our fixed training budget of natural language tokens, we aim for our models to be as comparable as those of \citet{hu-etal-2025-circuits} in terms of token-to-parameter ratio (\cref{tab:token-to-param}). Importantly, we note that all ratio values in both our and \citet{hu-etal-2025-circuits}'s experiments cannot be considered Chinchilla-optimal \citep{hoffmann2022trainingcomputeoptimallargelanguage}, which proposes the optimal value of roughly 20-25 tokens (or even more) per parameter found in recent open-weight LLMs \citep{grattafiori2024llama3herdmodels,gemmateam2025gemma3technicalreport,deepseekai2026deepseekv4}. Nonetheless, we see that our training and evaluation loss curves are stable in the last 3,000 training steps (e.g., in \cref{fig:eval_danish_481M_GT,subsec:A-4}), which suggests that our models are sufficiently trained and comparable to those of \citet{hu-etal-2025-circuits}. Crucially, we see little to no chance that the setups (baseline vs. pretrained) would change order with more training.

\begin{table}[h!]
\centering
\footnotesize
 
\begin{tabular}{|l|r| r| r| r|}
\hline
 & \textbf{Model} & \textbf{Size} & \textbf{Tokens} & \textbf{Ratio}\\ \hline \hline
\citet{hu-etal-2025-circuits} & Pythia & 160M & 655M & $\approx 4:1$ \\
 & Pythia & 1B & 1.63B & $1.63:1$ \\ \hline
Ours & \texttt{154M+LT} &253M & 655M & $\approx 2.6:1$ \\
(Llama-3-based & \texttt{154M+GT} &355M & 655M & $\approx 1.8:1$ \\
models) & \texttt{308M+LT} &440M & 655M & $\approx 1.5:1$ \\
 &  \texttt{308M+GT} &577M & 655M & $\approx 1.1:1$ \\
 &  \texttt{481M+LT} &678M & 655M & $\approx 0.97:1$ \\ 
 & \texttt{481M+GT} &884M & 655M & $\approx 0.74:1$ \\ \hline
\end{tabular}
\caption{Comparison of token-to-parameter ratios 
}
\label{tab:token-to-param}
\vspace{-12pt}
\end{table}

An important difference for our artificial language presentation is that \texttt{GT} splits
each number into separate digits,\footnote{e.g., \texttt{``81 209''} $\rightarrow$ \texttt{[`8', `1', `\_', `2', `0', `9']}}
different from
\texttt{LT}.\footnote{e.g., \texttt{``81 209''} $\rightarrow$ \texttt{[`81', `Ġ', `209']}}

Each model is trained for one epoch on the full training dataset (see \cref{subsec:natural_languages}) with three approaches: (1) without pretraining (\texttt{Base}); (2) after being pretrained with $k$-Dyck; and (3) after being pretrained with $k$-Shuffle Dyck. For (2) and (3), $k=64$ or $128$. For each setup, we collect three training runs with three different seed choices.
Following \citet{hu-etal-2025-circuits}, we quantify the improvement by token efficiency gain, which reflects how many training tokens sooner the model pretrained on the artificial language has reached the same validation loss as the corresponding \texttt{Base} model. If the pretrained model performs worse, we negate the value. 
Our exact calculation is provided in \cref{subsec:A-3}. Details of hyper-parameters used in our training are described in \cref{app:A}.

\subsection{Results}

\begin{figure}[t]
    \centering
    \includegraphics[width=1\linewidth]{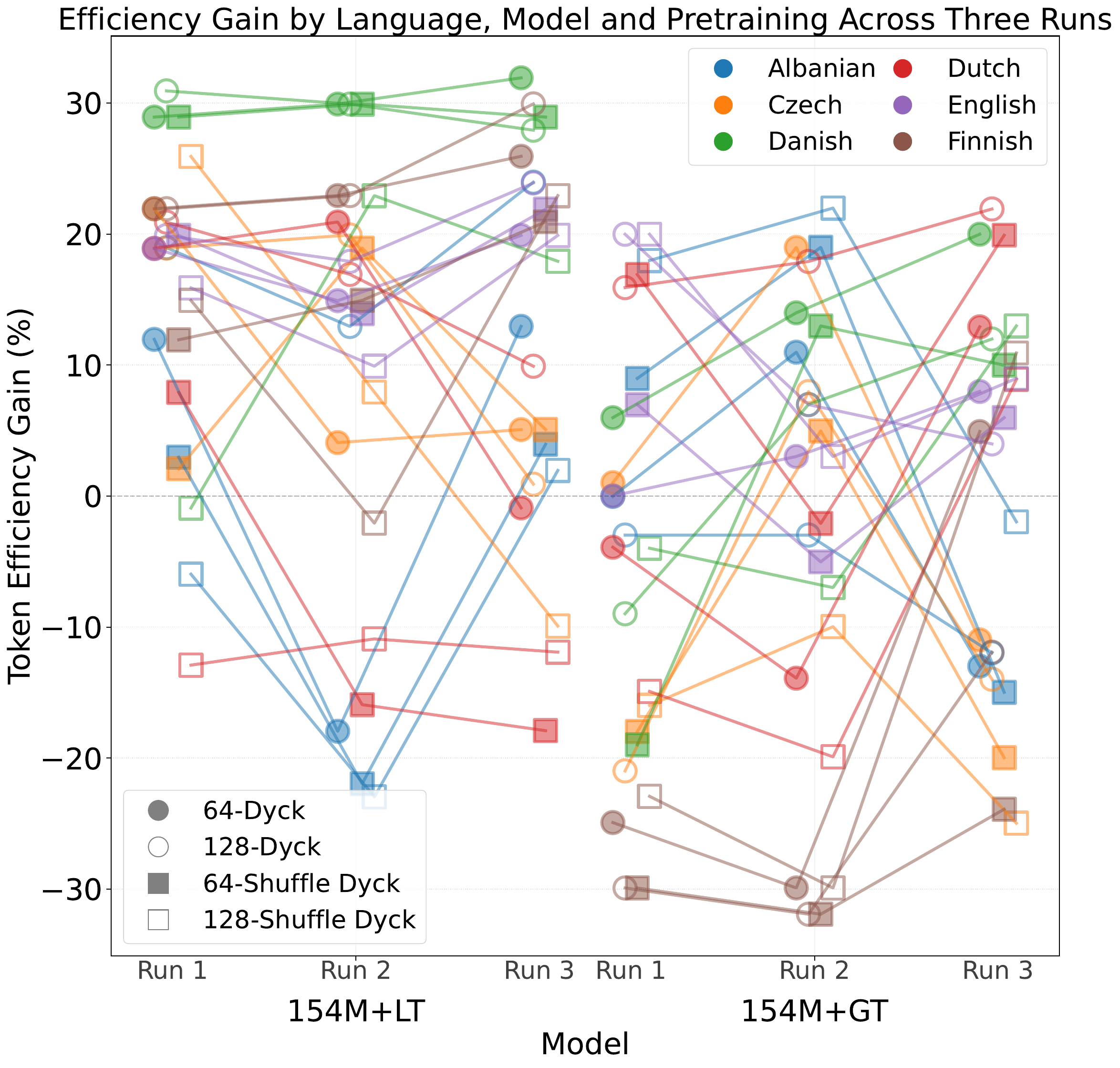}
    \vspace{-5pt}
    \caption{Average token efficiency gains/losses by language of \texttt{154M} models. The lines connect the identical setups across the three runs; horizontal lines thus indicate stable behavior, while lines crossing the x-axis indicate highly unstable cases where some training runs lead to a gain and some to a loss.
    }
    \label{fig:154M_3runs}
    \vspace{-10pt}
\end{figure}

The results across three runs reveal unexpectedly high variance within each identical setup,
observed both in the smallest (\cref{fig:154M_3runs}) 
and the bigger models (see \cref{fig:full_runs_comparison,fig:full_runs_comparison_by_lang} in \cref{app:B}).

\begin{table*}[t]
\centering
\begin{tabular}{|l||cccccccc||ccc|}
\hline
\multicolumn{1}{|c||}{\multirow{2}{*}{\makecell{\textbf{Natural} \\ \textbf{Language}}}} &
  \multicolumn{8}{c||}{\textbf{Pretraining Language}} &
  \multicolumn{3}{c|}{\textbf{Summary}} \\ \cline{2-12} 
\multicolumn{1}{|c||}{} &
  \multicolumn{2}{c|}{\makecell{\textbf{64-} \\ \textbf{Dyck}}} &
  \multicolumn{2}{c|}{\makecell{\textbf{128-} \\ \textbf{Dyck}}} &
  \multicolumn{2}{c|}{\makecell{\textbf{64-} \\ \textbf{Shuff} \\ \textbf{Dyck}}} &
  \multicolumn{2}{c||}{\makecell{\textbf{128-} \\ \textbf{Shuff} \\ \textbf{Dyck}}} &
  \multicolumn{1}{c|}{\texttt{LT}} &
  \multicolumn{1}{c|}{\texttt{GT}} &
  \textbf{Overall} \\ \hline \hline
Albanian &
  \multicolumn{1}{c|}{$\times$} &
  \multicolumn{1}{c|}{$\times$} &
  \multicolumn{1}{c|}{$+$} &
  \multicolumn{1}{c|}{$-$} &
  \multicolumn{1}{c|}{$\times$} &
  \multicolumn{1}{c|}{$\times$} &
  \multicolumn{1}{c|}{$\times$} &
  $\times$ &
  \multicolumn{1}{c|}{1$+$  3$\times$  0$-$} &
  \multicolumn{1}{c|}{0$+$  3$\times$  1$-$} &
  1$+$  6$\times$  1$-$ \\ \hline
Czech &
  \multicolumn{1}{c|}{$+$} &
  \multicolumn{1}{c|}{$\times$} &
  \multicolumn{1}{c|}{$+$} &
  \multicolumn{1}{c|}{$\times$} &
  \multicolumn{1}{c|}{$+$} &
  \multicolumn{1}{c|}{$\times$} &
  \multicolumn{1}{c|}{$\times$} &
  $-$ &
  \multicolumn{1}{c|}{3$+$  1$\times$  0$-$} &
  \multicolumn{1}{c|}{0$+$  3$\times$  1$-$} &
  3$+$  4$\times$  1$-$ \\ \hline
Danish &
  \multicolumn{1}{c|}{$\times$} &
  \multicolumn{1}{c|}{$+$} &
  \multicolumn{1}{c|}{$+$} &
  \multicolumn{1}{c|}{$\times$} &
  \multicolumn{1}{c|}{$+$} &
  \multicolumn{1}{c|}{$\times$} &
  \multicolumn{1}{c|}{$\times$} &
  $\times$ &
  \multicolumn{1}{c|}{2$+$  2$\times$  0$-$} &
  \multicolumn{1}{c|}{1$+$  3$\times$  0$-$} &
  3$+$  5$\times$  0$-$ \\ \hline
Dutch &
  \multicolumn{1}{c|}{$+$} &
  \multicolumn{1}{c|}{$\times$} &
  \multicolumn{1}{c|}{$+$} &
  \multicolumn{1}{c|}{$+$} &
  \multicolumn{1}{c|}{$\times$} &
  \multicolumn{1}{c|}{$\times$} &
  \multicolumn{1}{c|}{$-$} &
  $-$ &
  \multicolumn{1}{c|}{2$+$  1$\times$  1$-$} &
  \multicolumn{1}{c|}{1$+$  2$\times$  1$-$} &
  3$+$ 3$\times$ 2$-$ \\ \hline
English &
  \multicolumn{1}{c|}{$+$} &
  \multicolumn{1}{c|}{$+$} &
  \multicolumn{1}{c|}{$+$} &
  \multicolumn{1}{c|}{$+$} &
  \multicolumn{1}{c|}{$+$} &
  \multicolumn{1}{c|}{$\times$} &
  \multicolumn{1}{c|}{$+$} &
  $+$ &
  \multicolumn{1}{c|}{4$+$  0$\times$  0$-$} &
  \multicolumn{1}{c|}{3$+$  1$\times$  0$-$} &
  7$+$  1$\times$  0$-$ \\ \hline
Finnish &
  \multicolumn{1}{c|}{$+$} &
  \multicolumn{1}{c|}{$\times$} &
  \multicolumn{1}{c|}{$+$} &
  \multicolumn{1}{c|}{$-$} &
  \multicolumn{1}{c|}{$+$} &
  \multicolumn{1}{c|}{$-$} &
  \multicolumn{1}{c|}{$\times$} &
  $\times$ &
  \multicolumn{1}{c|}{3$+$  1$\times$  0$-$} &
  \multicolumn{1}{c|}{0$+$  2$\times$  2$-$} &
  3$+$  3$\times$  2$-$ \\ \hline
\end{tabular}
\caption{
Consistency of the results of pretraining with the \texttt{154M} models. We only indicate if the setup lead to a \ul{consistent gain} ($+$) over the \texttt{Base} model in all three runs, to a \ul{consistent loss} ($-$), or to a \ul{mixed} result ($\times$) where some of the runs benefited and some lost.
In each pretraining language group, the left and right columns represent the result with \texttt{LT} and \texttt{GT}, respectively. The summaries report the number of consistent gains, mixed outcomes and losses across the respective setting.}
\label{tab:efficiency_by_pretraining} 
\vspace{-10pt}
\end{table*}

\begin{figure}[h!]
    \centering
    \includegraphics[width=\linewidth]{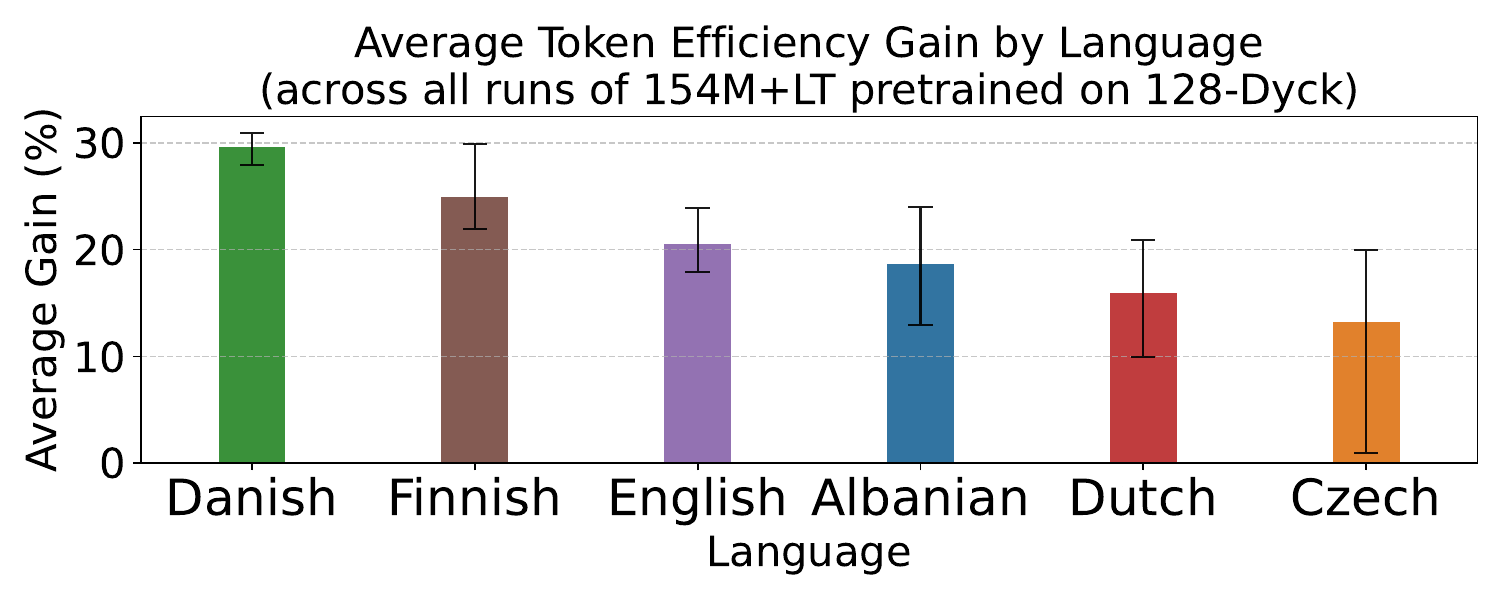}
    \vspace{-15pt}
    \caption{Average token efficiency gains by language (for \texttt{154M+LT} pretrained on 128-Dyck)}
    \label{fig:average_gains_best_setup}
    \vspace{-10pt}
\end{figure}

However, in one specific setup, namely, \texttt{154M+LT} pretrained on 128-Dyck, pretraining does bring consistent improvement (\cref{fig:average_gains_best_setup,tab:efficiency_by_pretraining}).
In addition, our experimental results with English shown in \cref{tab:efficiency_by_pretraining} suggests that pretraining the smallest \texttt{154M} models with most artificial languages (barring only 64-Shuff Dyck) also leads to constant gain; therefore, we can safely corroborate the findings of previous studies for English (see \cref{sec:related_work}).

Notably, models with \texttt{LT} on average have higher gains than those with \texttt{GT} (\cref{fig:tokenizers_comparison}). For the former, $k$-Dyck usually outperforms $k$-Shuffle Dyck in terms of gains, while the pattern is not that clear for the latter models 
(see \cref{fig:full_runs_comparison} in \cref{app:B}). 

\begin{figure}[h!]
    \centering
    \includegraphics[width=\linewidth]{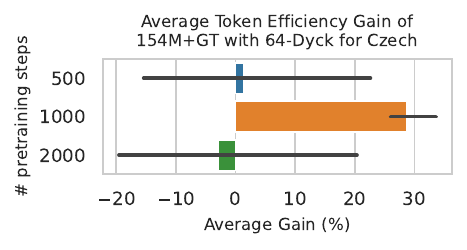}
    \vspace{-15pt}
    \caption{Average token efficiency gains of \texttt{154M+GT} pretrained on 64-Dyck for Czech, with three different pretraining checkpoints. Right is better.}
    \label{fig:154M_GT_czech}
    \vspace{-15pt}
\end{figure}

We also conduct an additional experiment with one of the least stable setups---the \texttt{154M+GT} model pretrained on 64-Dyck and subsequently trained on Czech---which we pretrain with 500, 1,000, and 2,000 steps, three runs per value. \cref{fig:154M_GT_czech} suggests that pretraining with 1,000 steps could be optimal.

\section{Language Specific Analysis}

Might there be any correlation between the gain and the linguistic properties? Due to observed inconsistencies, we limit the analysis to \texttt{154M} models and conduct significance testing. All definitions of the linguistic metrics used, along with all details of our language-specific analysis are provided in \cref{app:C,app:D}, respectively.

We observe that only the \texttt{154M+GT} models show statistically significant correlations stable across runs and pretraining languages.
For these setups, positive correlation can generally be seen with \textit{Mean Children}, \textit{Average Perplexity},
and \textit{Average Sentence Length}. Whereas, the level of \textit{Morphological Richness}, \textit{Vocabulary Size}, and \textit{Max. Number of Children} corresponds to lower efficiency gains or losses (see \cref{fig:average_correlation_GT_eval,fig:average_correlation_GT_train} in \cref{app:D}).

\begin{figure}[h!]
    \centering
    \includegraphics[width=\linewidth]{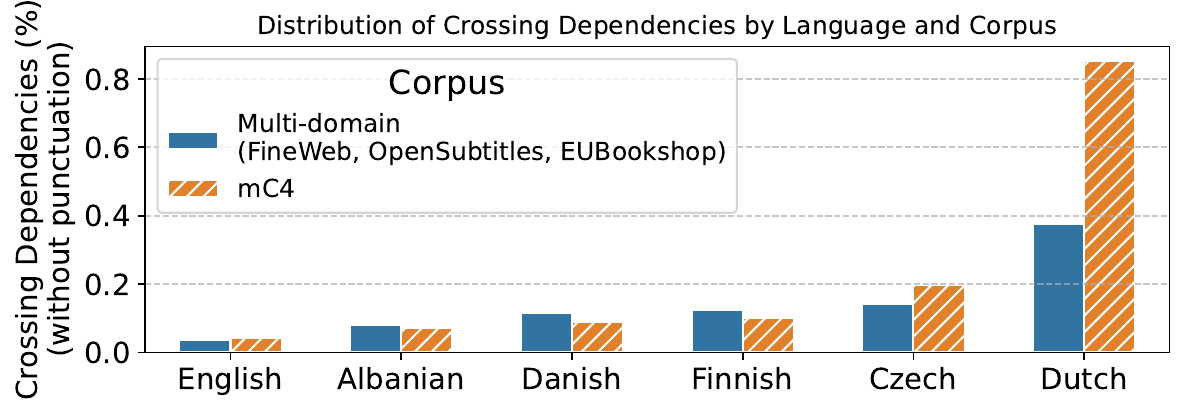}
    \caption{Distribution of crossing dependencies by language and corpus, sorted by the percentage value}
    \label{fig:crossing_dep_distribution}
    \vspace{-10pt}
\end{figure}

{\RaggedRight \textbf{Crossing Dependencies.}}
Our primary hypothesis was that models pretrained on $k$-Shuffle Dyck would benefit from data with high \textit{Percentage of Crossing Dependencies}. We expected the gains to correspond to the distribution of crossing dependencies 
(\cref{fig:crossing_dep_distribution}).
However, we
do not detect neither stable across runs nor average positive statistically significant
correlations between gains of models pretrained on $k$-Shuffle Dyck and 
\textit{Percentage of Crossing Dependencies}
in either data split. 
(see \cref{fig:average_correlation_eval_shuff,fig:average_correlation_train_shuff} in \cref{app:D}).

\section{Conclusion}
\label{sec:6}
In this work, we investigate the benefit of pretraining LLMs with artificial languages by extending the experiments to five other natural languages aside from English. Our results suggest that the approach may work for other languages than English too. However, it is highly sensitive to the exact setup, i.e., artificial language, tokenizer choice, and the amount of pretraining.
Often, training runs differing only in the random seed lead to different gains or losses. We thus stress the need to validate experimental results by multiple runs. With the pretraining approach, we observe the most stability with small models using the Llama tokenizer and pretrained on 128-Dyck;
while all other setups produce mixed results. Finally, we find almost no relation of the gains to over 20 linguistic properties.


\section*{Limitations}
The work presented in this paper faces several limitations that restricted us from having more comprehensive results.

First, one noticeable issue concerns the easily understandable lack of a large-scale, multi-parallel dataset in multiple languages, both in training and evaluation. Having this would 
allow us to have better and fairer comparisons between models. The currently available datasets are mostly from the Internet, which might not reflect the highest quality in the language modeling task.

Second, due to the sheer scale of the experiments, we can only train each model with 10,000 training steps (which corresponds to approximately 655M tokens). In addition, on the technical side, we fail to conduct more experiments with a wider range and combination of training hyper-parameters. Even though the resulting models successfully converged during training, we cannot claim that our chosen set of training hyper-parameters is the best one.

Third, we acknowledge that with just 655M training tokens, our models could not be considered Chinchilla-optimal \citep{hoffmann2022trainingcomputeoptimallargelanguage}, and should be viewed as ``undertrained'' compared to any production-grade LLMs.
We nevertheless believe that the presented learning curves are stable enough to convincingly identify if the pre-pretraining has helped or harmed in that particular run.

Fourth, we only conduct experiments with six languages from the European region, which share the same Latin script. As a result, we can only claim the analyses for the examined languages, and this study should not be viewed as a proof that the method works for most, if not all, languages. We leave additional languages and writing scripts for future research.

\section*{Acknowledgments}
This research has been supported by the OpenEuroLLM project, funded by the EC Digital Europe Programme under grant agreement No. 101195233, and partially supported by SVV project number 260 821.


\bibliography{custom}



\appendix
\section{Details of Experiment Setups}
\label{app:A}
\subsection{Model hyper-parameters}
We take the model configuration of the published 
\texttt{meta-llama/Llama-3.2-1B}
checkpoint, then modify four hyper-parameters as described in \cref{tab:config_hyper_params}. For each setup, we collect three training runs with three different seed choices of $17$, $42$, and $3407$.

\begin{table}[h!]
\centering
\begin{tabular}{|l|r|r|r|r|}
\hline
\textbf{Config} & \multicolumn{1}{r|}{0} & \multicolumn{1}{r|}{1} & \multicolumn{1}{r|}{2} \\ \hline \hline
\texttt{num\_hidden\_layers}  & 8   & 12    & 16    \\ \hline
\texttt{num\_attn\_heads}     & 8   & 8     & 16    \\ \hline
\texttt{head\_dim}       & 16  & 16    & 32    \\ \hline
\texttt{hidden\_size}    & 768 & 1,024 & 1,536 \\ \hline
\end{tabular}
\caption{Details of models' hyper-parameters.}
\label{tab:config_hyper_params}
\end{table}

\subsection{Training hyper-parameters}
\cref{tab:train_hyper_params} details chosen training hyper-parameters.
\begin{table}[h!]
\vspace{-7pt}
\centering
\begin{tabular}{|l|l|}
\hline
\multicolumn{1}{|c|}{\textbf{Training hyper-parameters}} & \multicolumn{1}{c|}{\textbf{Value}} \\ \hline \hline
Effective batch size                                     & 32                                  \\ \hline
Learning rate                                            & 5e-4                                \\ \hline
Optimizer                                                & AdamW                               \\ \hline
LR Scheduler                                             & cosine                              \\ \hline
Min. learning rate                                       & 0.1                                 \\ \hline
Warmup steps                                             & 1,000                               \\ \hline
Weight decay                                             & 0.1                                 \\ \hline
Max. grad. norm.                                         & 1.0                                 \\ \hline
\end{tabular}
\caption{Training hyper-parameters}
\label{tab:train_hyper_params}
\end{table}

\subsection{Algorithm to Calculate Token Efficiency Gain}
\label{subsec:A-3}
The algorithm to calculate the token efficiency gain is described in \cref{alg:efficiency_gain}.

\begin{algorithm*}[h!]
\SetAlgoLined
\DontPrintSemicolon
\caption{Calculate Token Efficiency Gain}
\label{alg:efficiency_gain}
\KwIn{
    Sequence of the baseline model's validation losses $\textcolor{red}{B}$, 
    Sequence of the pretrained model's validation
    losses $\textcolor{blue}{P}$
}
\KwOut{Token Efficiency Gain (\%)}
\;
$PrePretrainedTokens \leftarrow$ Total number of tokens in the pre-pretraining step\;
$\textcolor{red}{BaselineTokens} \leftarrow$ Total number of tokens used by the baseline model\;
$TargetLoss \leftarrow$ Final value in $\textcolor{red}{B}$\;
\;
\If{$i \geq TargetLoss$ $\forall i \in \textcolor{blue}{P}$}{
    $ Gain \leftarrow \text{Calculate Token Efficiency Gain}(\textcolor{blue}{P}, \textcolor{red}{B})$\;
    \Return $-Gain$\;
}
\Else{
    $n \leftarrow length(\textcolor{blue}{P})$\;
    
    $k \leftarrow \min\{j\in \{0,\dots,n-1\} \text{ such that}$
    $\forall i ,j\le i< n: \textcolor{blue}{P}[i]\le TargetLoss\}$\;
    
    
    $\textcolor{blue}{PretrainedTokens} \leftarrow$ Number of tokens used by the pretrained model at index $k$\;
    $\textcolor{blue}{PretrainedTokens} \leftarrow \textcolor{blue}{PretrainedTokens} + PrePretrainedTokens$\;
    $Gain \leftarrow (1 - \frac{\textcolor{blue}{PretrainedTokens}}{\textcolor{red}{BaselineTokens}}) \times 100$\;
    \Return $Gain$\;
}
\end{algorithm*}

\subsection{Training and Evaluation Losses}
\label{subsec:A-4}
\begin{figure*}[h!]
    \centering
    \begin{subfigure}[b]{\textwidth}
        \includegraphics[width=\linewidth]{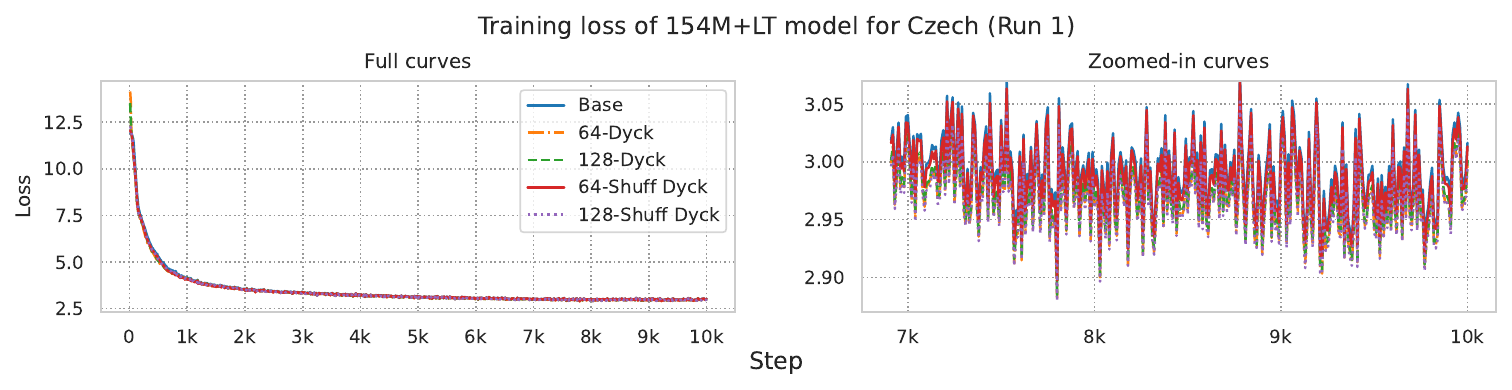}
        \caption{Training Loss}
        \label{fig:czech_154-LT_train}
    \end{subfigure}

    \begin{subfigure}[b]{\textwidth}
        \includegraphics[width=\linewidth]{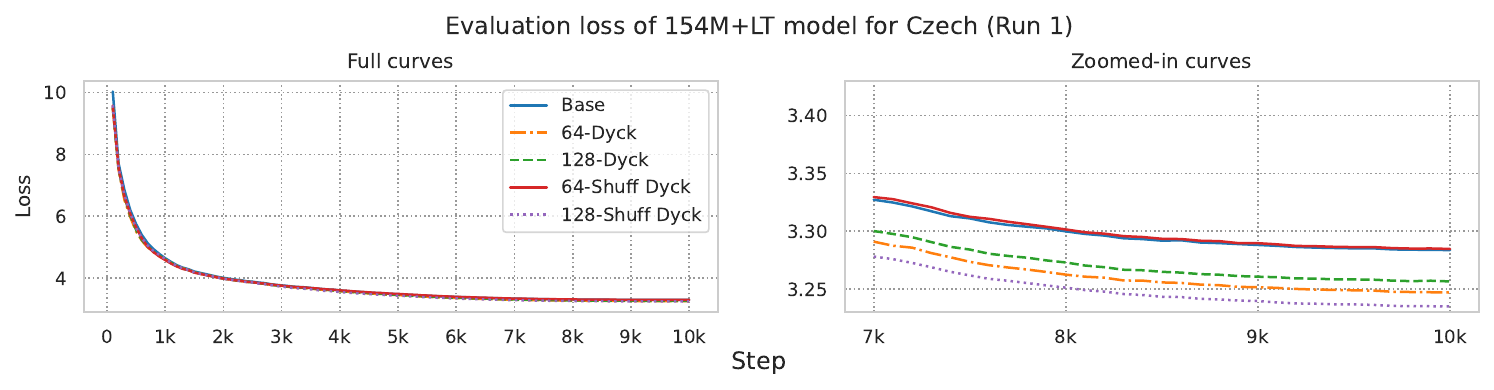}
        \caption{Evaluation Loss}
        \label{fig:czech_154-LT_eval}
    \end{subfigure}
    \caption{
    Loss curves of the \texttt{154M+LT} model for Czech. The right-side figure is the zoomed-in version of the left one, from step 7,000 to 10,000.}
    \label{fig:czech_154-LT}
\end{figure*}

\begin{figure*}[h!]
    \centering
    \begin{subfigure}[b]{\textwidth}
        \includegraphics[width=\linewidth]{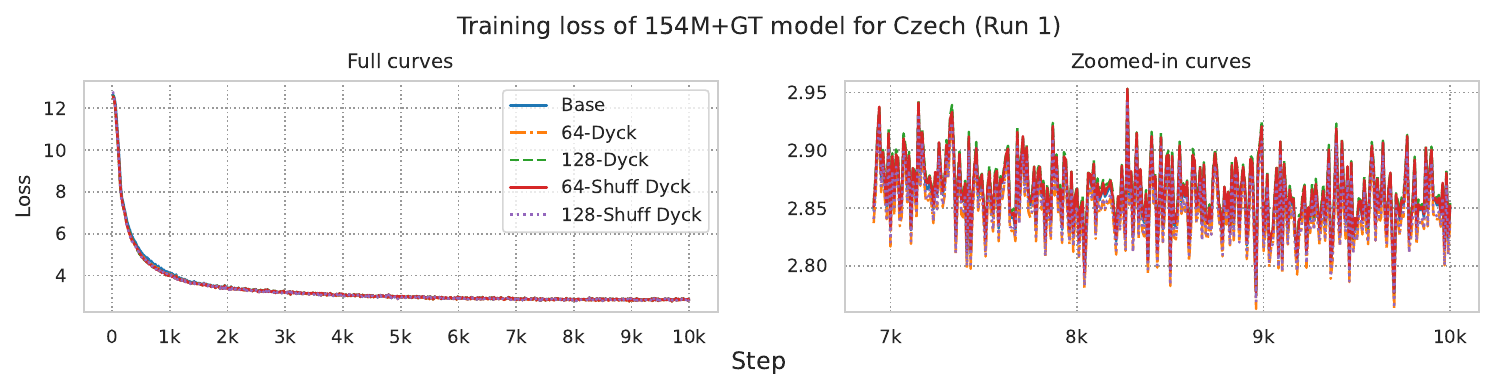}
        \caption{Training Loss}
        \label{fig:czech_154-GT_train}
    \end{subfigure}

    \begin{subfigure}[b]{\textwidth}
        \includegraphics[width=\linewidth]{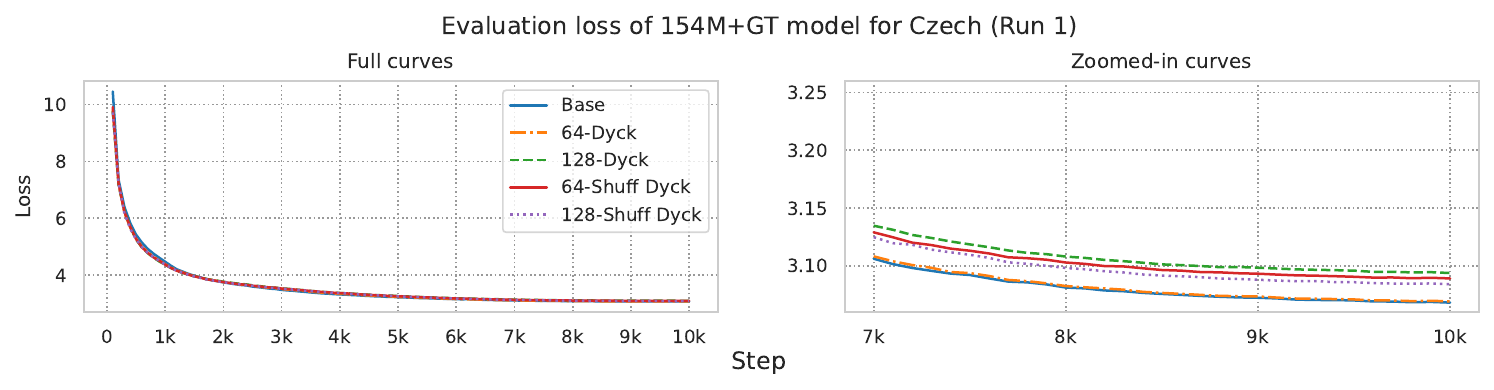}
        \caption{Evaluation Loss}
        \label{fig:czech_154-GT_eval}
    \end{subfigure}
    \caption{Loss curves of the \texttt{154M+GT} model for Czech. The right-side figure is the zoomed-in version of the left one, from step 7,000 to 10,000.}
    \label{fig:czech_154-GT}
\end{figure*}

\begin{figure*}[h!]
    \centering
    \begin{subfigure}[b]{\textwidth}
        \includegraphics[width=\linewidth]{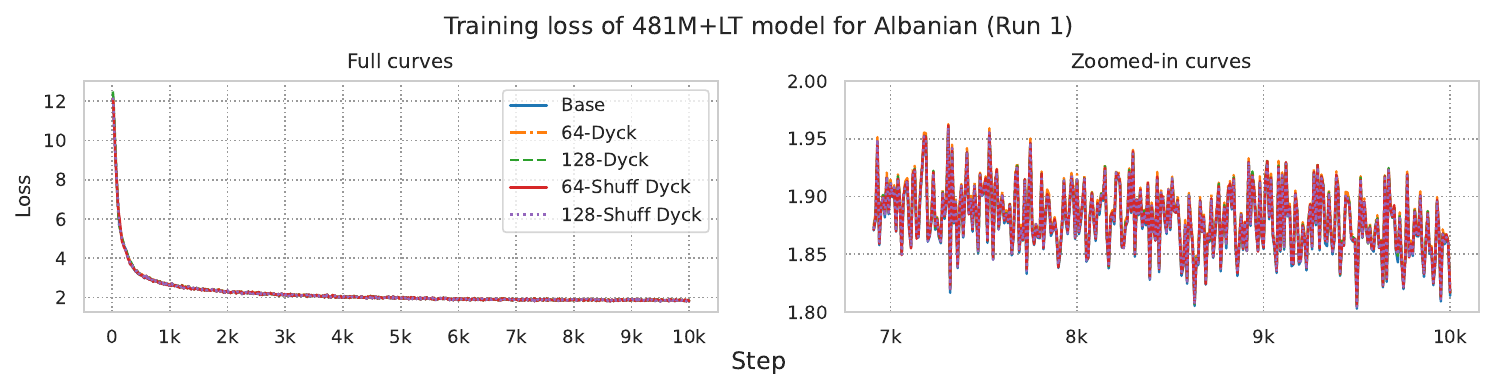}
        \caption{Training Loss}
        \label{fig:albanian_481-LT_train}
    \end{subfigure}

    \begin{subfigure}[b]{\textwidth}
        \includegraphics[width=\linewidth]{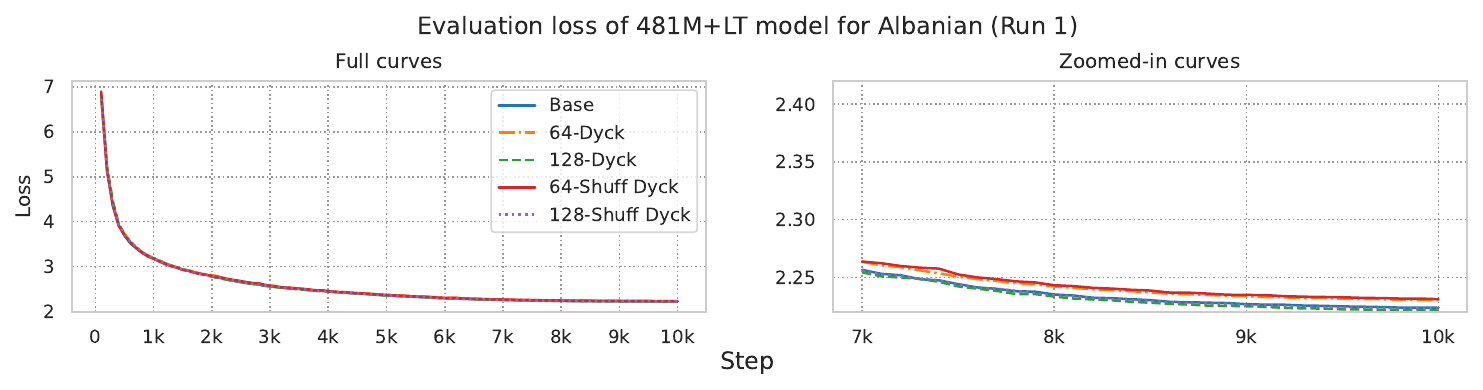}
        \caption{Evaluation Loss}
        \label{fig:albanian_481-LT_eval}
    \end{subfigure}
    \caption{Loss curves of the \texttt{481M+LT} model for Albanian. The right-side figure is the zoomed-in version of the left one, from step 7,000 to 10,000.}
    \label{fig:albanian_481-LT}
\end{figure*}

\begin{figure*}[h!]
    \centering
    \begin{subfigure}[b]{\textwidth}
        \includegraphics[width=\linewidth]{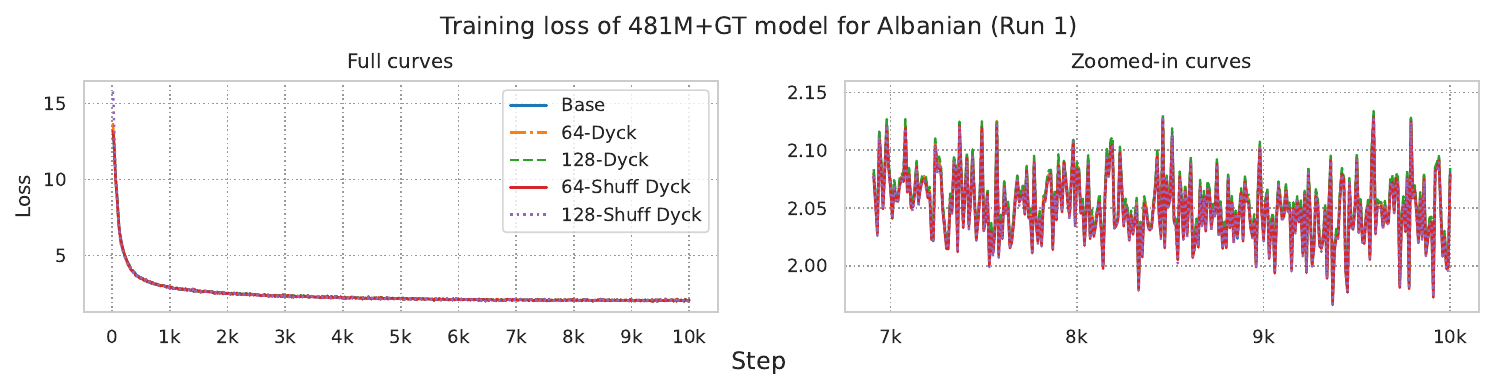}
        \caption{Training Loss}
        \label{fig:albanian_481-GT_train}
    \end{subfigure}

    \begin{subfigure}[b]{\textwidth}
        \includegraphics[width=\linewidth]{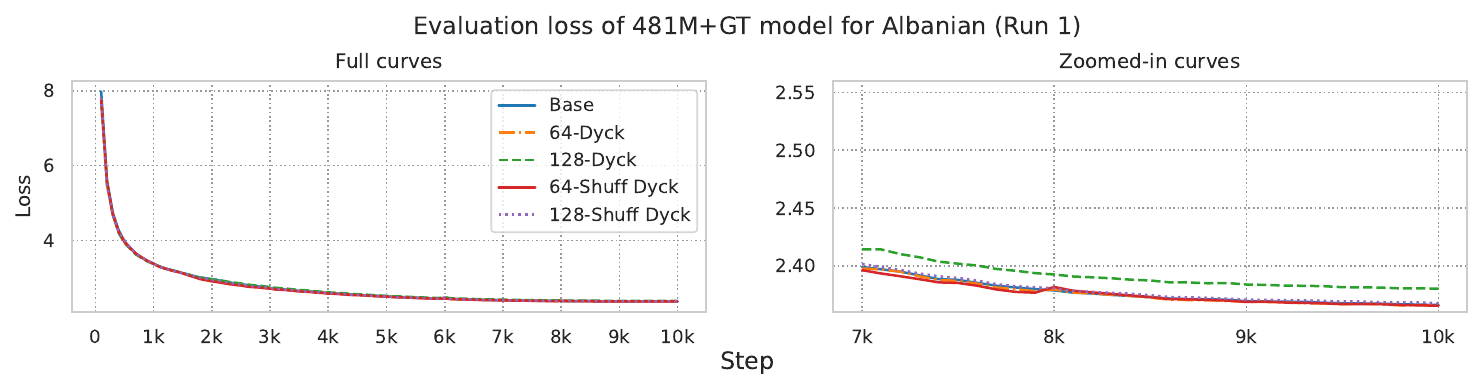}
        \caption{Evaluation Loss}
        \label{fig:albanian_481-GT_eval}
    \end{subfigure}
    \caption{Loss curves of the \texttt{481M+GT} model for Albanian. The right-side figure is the zoomed-in version of the left one, from step 7,000 to 10,000.}
    \label{fig:albanian_481-GT}
\end{figure*}

We provide some figures describing the loss curves within our experiments.
\cref{fig:czech_154-LT,fig:czech_154-GT} show the training and evaluation loss curves of the \texttt{154M+\{LT,GT\}} models for Czech. \cref{fig:albanian_481-LT,fig:albanian_481-GT} describe the training and evaluation loss curves of the \texttt{154M+\{LT,GT\}} models for Albanian. 

\section{Details of Token Efficiency Gain} \label{app:B}

\autoref{tab:efficiency_by_pretraining} describes the consistency of evaluation results for the same model (\texttt{154M}) with different tokenizers (Llama - \texttt{LT} or Gemma - \texttt{GT}) across all pretraining and natural languages.
Symbols  `$+$', `$-$', and `$\times$' denote \ul{consistent gain}, \ul{consistent loss}, and \ul{inconsistency} across three runs (i.e., some of the runs resulted in efficiency gain, while others in loss or negligible (near-zero) gain), respectively. 



\begin{figure*}[h!]
    \centering
    \includegraphics[width=\linewidth]{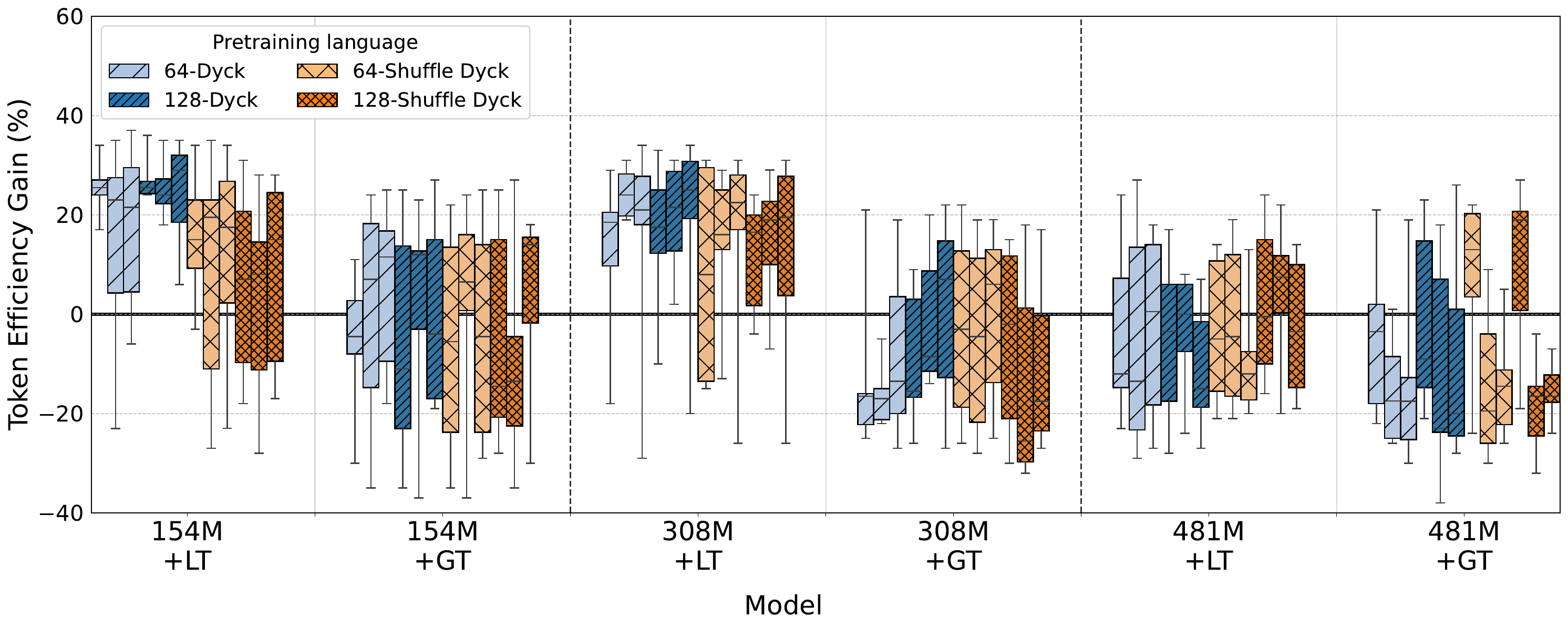}
    \caption{Efficiency gain across all models, pretraining, and natural languages. }
    \label{fig:full_runs_comparison}
\end{figure*}
\begin{figure*}[h!]
    \centering
    \includegraphics[width=\linewidth]{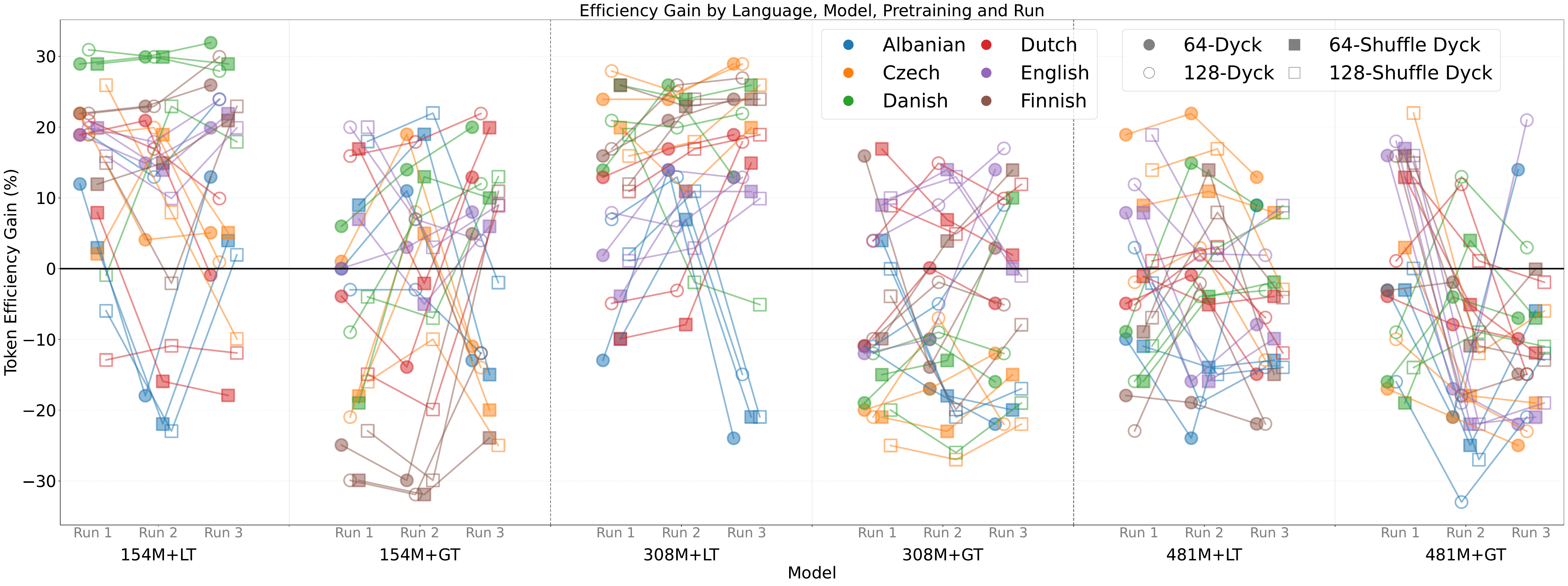}
    \caption{Efficiency gain across all models, pretraining, and natural languages (with language-specific details). The lines connect the identical setups across the three runs; horizontal lines thus indicate stable behavior, while lines crossing the x-axis indicate highly unstable cases where some training runs lead to a gain and some to a loss.}
\label{fig:full_runs_comparison_by_lang}
\end{figure*}

\cref{fig:full_runs_comparison,fig:full_runs_comparison_by_lang} describe how the token efficiency gain differs between all model setups, runs, pretraining, and natural languages.

\autoref{tab:token_efficiency_full} provides performance gain (or loss) results for all experiments.

\clearpage
\clearpage
\begin{small}
\setlength{\tabcolsep}{2.5pt}
\tablehead{\hline \textbf{Model} & \textbf{Language} & \textbf{Gain \% $\uparrow$} \\ \hline \hline}
\topcaption{Token efficiency gain across experiments} \label{tab:token_efficiency_full}
\begin{supertabular}{|l|c|r|}
481M+GT\_128-dyck\_run2 & Albanian & -32.980 \\
154M+GT\_128-dyck\_run2 & Finnish & -31.915 \\
154M+GT\_64-shuff-dyck\_run2 & Finnish & -31.915 \\
154M+GT\_128-dyck\_run1 & Finnish & -29.915 \\
154M+GT\_64-shuff-dyck\_run1 & Finnish & -29.915 \\
154M+GT\_128-shuff-dyck\_run2 & Finnish & -29.911 \\
154M+GT\_64-dyck\_run2 & Finnish & -29.911 \\
481M+GT\_128-shuff-dyck\_run2 & Albanian & -26.983 \\
308M+GT\_128-shuff-dyck\_run2 & Czech & -26.982 \\
308M+GT\_128-shuff-dyck\_run2 & Danish & -25.985 \\
308M+GT\_128-shuff-dyck\_run1 & Czech & -24.981 \\
481M+GT\_64-dyck\_run3 & Czech & -24.974 \\
481M+GT\_64-shuff-dyck\_run2 & Albanian & -24.974 \\
154M+GT\_128-shuff-dyck\_run3 & Czech & -24.973 \\
154M+GT\_64-dyck\_run1 & Finnish & -24.919 \\
308M+LT\_64-dyck\_run3 & Albanian & -23.966 \\
481M+LT\_64-dyck\_run2 & Albanian & -23.951 \\
154M+GT\_64-shuff-dyck\_run3 & Finnish & -23.904 \\
308M+GT\_64-shuff-dyck\_run2 & Czech & -22.982 \\
481M+GT\_128-dyck\_run3 & Czech & -22.982 \\
154M+LT\_128-shuff-dyck\_run2 & Albanian & -22.956 \\
481M+LT\_128-dyck\_run1 & Finnish & -22.931 \\
154M+GT\_128-shuff-dyck\_run1 & Finnish & -22.918 \\
481M+GT\_64-dyck\_run3 & English & -21.977 \\
481M+GT\_128-shuff-dyck\_run2 & English & -21.977 \\
481M+GT\_64-shuff-dyck\_run2 & English & -21.977 \\
308M+GT\_128-dyck\_run3 & Czech & -21.970 \\
308M+GT\_128-shuff-dyck\_run3 & Czech & -21.970 \\
308M+GT\_64-dyck\_run3 & Albanian & -21.968 \\
154M+LT\_64-shuff-dyck\_run2 & Albanian & -21.956 \\
481M+LT\_128-dyck\_run3 & Finnish & -21.936 \\
481M+LT\_64-dyck\_run3 & Finnish & -21.936 \\
481M+GT\_64-shuff-dyck\_run3 & English & -20.982 \\
308M+GT\_128-dyck\_run1 & Czech & -20.978 \\
154M+GT\_128-dyck\_run1 & Czech & -20.978 \\
308M+GT\_64-shuff-dyck\_run1 & Czech & -20.978 \\
481M+GT\_64-dyck\_run2 & Czech & -20.974 \\
308M+GT\_128-shuff-dyck\_run2 & Albanian & -20.971 \\
481M+GT\_128-dyck\_run3 & Albanian & -20.971 \\
481M+GT\_64-dyck\_run2 & Albanian & -20.971 \\
308M+LT\_128-shuff-dyck\_run3 & Albanian & -20.950 \\
308M+LT\_64-shuff-dyck\_run3 & Albanian & -20.950 \\
308M+GT\_128-shuff-dyck\_run1 & Danish & -19.980 \\
308M+GT\_64-dyck\_run1 & Czech & -19.979 \\
154M+GT\_64-shuff-dyck\_run3 & Czech & -19.976 \\
308M+GT\_64-shuff-dyck\_run3 & Albanian & -19.974 \\
308M+GT\_128-shuff-dyck\_run2 & Finnish & -19.906 \\
154M+GT\_128-shuff-dyck\_run2 & Dutch & -19.902 \\
308M+GT\_128-shuff-dyck\_run3 & Danish & -18.985 \\
481M+GT\_64-shuff-dyck\_run3 & Czech & -18.981 \\
308M+GT\_64-dyck\_run1 & Danish & -18.980 \\
481M+GT\_64-shuff-dyck\_run1 & Danish & -18.980 \\
154M+GT\_64-shuff-dyck\_run1 & Danish & -18.980 \\
481M+GT\_128-shuff-dyck\_run3 & English & -18.976 \\
481M+GT\_128-dyck\_run2 & English & -18.976 \\
481M+LT\_128-dyck\_run2 & Albanian & -18.949 \\
481M+LT\_64-dyck\_run2 & Finnish & -18.929 \\
308M+GT\_64-shuff-dyck\_run2 & Albanian & -17.980 \\
154M+GT\_64-shuff-dyck\_run1 & Czech & -17.978 \\
481M+GT\_64-shuff-dyck\_run2 & Czech & -17.969 \\
481M+GT\_128-dyck\_run2 & Czech & -17.969 \\
154M+LT\_64-dyck\_run2 & Albanian & -17.955 \\
481M+LT\_64-dyck\_run1 & Finnish & -17.925 \\
154M+LT\_64-shuff-dyck\_run3 & Dutch & -17.908 \\
481M+GT\_128-dyck\_run2 & Finnish & -17.902 \\
308M+GT\_64-dyck\_run2 & Albanian & -16.983 \\
481M+GT\_64-dyck\_run1 & Czech & -16.979 \\
308M+GT\_64-dyck\_run2 & Czech & -16.973 \\
481M+GT\_64-dyck\_run2 & English & -16.970 \\
308M+GT\_128-shuff-dyck\_run3 & Albanian & -16.968 \\
308M+GT\_64-dyck\_run3 & Danish & -15.982 \\
481M+GT\_64-dyck\_run1 & Danish & -15.981 \\
154M+GT\_128-shuff-dyck\_run1 & Czech & -15.979 \\
481M+GT\_128-dyck\_run1 & Albanian & -15.977 \\
154M+LT\_64-shuff-dyck\_run2 & Dutch & -15.919 \\
481M+LT\_128-dyck\_run1 & Danish & -15.916 \\
481M+LT\_64-shuff-dyck\_run1 & Danish & -15.916 \\
481M+LT\_64-shuff-dyck\_run2 & English & -15.915 \\
481M+LT\_64-dyck\_run2 & English & -15.915 \\
308M+GT\_64-shuff-dyck\_run1 & Danish & -14.981 \\
308M+GT\_64-shuff-dyck\_run3 & Czech & -14.980 \\
154M+GT\_64-shuff-dyck\_run3 & Albanian & -14.975 \\
308M+LT\_128-dyck\_run3 & Albanian & -14.948 \\
481M+LT\_128-shuff-dyck\_run2 & Albanian & -14.948 \\
481M+LT\_64-shuff-dyck\_run3 & Finnish & -14.931 \\
481M+GT\_128-dyck\_run3 & Finnish & -14.916 \\
481M+GT\_64-dyck\_run3 & Finnish & -14.916 \\
481M+LT\_64-dyck\_run3 & Dutch & -14.912 \\
481M+GT\_128-dyck\_run3 & Dutch & -14.910 \\
154M+GT\_128-shuff-dyck\_run1 & Dutch & -14.909 \\
154M+GT\_128-dyck\_run3 & Czech & -13.982 \\
481M+GT\_128-shuff-dyck\_run1 & Danish & -13.980 \\
481M+LT\_64-shuff-dyck\_run2 & Albanian & -13.948 \\
481M+LT\_128-dyck\_run3 & Albanian & -13.948 \\
481M+LT\_128-shuff-dyck\_run3 & Albanian & -13.948 \\
308M+GT\_64-dyck\_run2 & Finnish & -13.909 \\
154M+GT\_64-dyck\_run2 & Dutch & -13.902 \\
154M+GT\_64-dyck\_run3 & Albanian & -12.981 \\
308M+GT\_64-shuff-dyck\_run2 & Danish & -12.980 \\
308M+LT\_64-dyck\_run1 & Albanian & -12.957 \\
481M+LT\_64-shuff-dyck\_run3 & Albanian & -12.947 \\
481M+LT\_128-dyck\_run3 & Czech & -12.920 \\
481M+GT\_128-shuff-dyck\_run3 & Finnish & -12.913 \\
154M+LT\_128-shuff-dyck\_run1 & Dutch & -12.904 \\
308M+GT\_128-dyck\_run3 & Danish & -11.984 \\
308M+GT\_64-dyck\_run1 & English & -11.983 \\
308M+GT\_128-dyck\_run1 & Danish & -11.981 \\
481M+GT\_128-shuff-dyck\_run2 & Czech & -11.976 \\
308M+GT\_64-dyck\_run3 & Czech & -11.976 \\
154M+GT\_128-dyck\_run3 & Albanian & -11.969 \\
481M+GT\_128-shuff-dyck\_run3 & Albanian & -11.968 \\
481M+LT\_128-shuff-dyck\_run3 & Dutch & -11.915 \\
154M+LT\_128-shuff-dyck\_run3 & Dutch & -11.914 \\
154M+GT\_128-dyck\_run3 & Finnish & -11.912 \\
481M+GT\_64-shuff-dyck\_run3 & Dutch & -11.910 \\
481M+GT\_128-shuff-dyck\_run3 & Danish & -10.988 \\
154M+GT\_64-dyck\_run3 & Czech & -10.978 \\
308M+GT\_128-dyck\_run1 & Albanian & -10.975 \\
308M+GT\_64-dyck\_run1 & Albanian & -10.975 \\
481M+LT\_64-shuff-dyck\_run1 & Albanian & -10.958 \\
481M+LT\_128-shuff-dyck\_run1 & Danish & -10.912 \\
481M+GT\_128-shuff-dyck\_run2 & Finnish & -10.911 \\
481M+GT\_64-shuff-dyck\_run2 & Finnish & -10.911 \\
154M+LT\_128-shuff-dyck\_run2 & Dutch & -10.903 \\
308M+GT\_64-dyck\_run1 & Dutch & -10.899 \\
308M+GT\_64-dyck\_run2 & Danish & -9.993 \\
481M+GT\_128-dyck\_run1 & Czech & -9.979 \\
308M+GT\_64-dyck\_run2 & English & -9.973 \\
154M+GT\_128-shuff-dyck\_run2 & Czech & -9.971 \\
481M+LT\_64-dyck\_run1 & Albanian & -9.958 \\
308M+LT\_64-shuff-dyck\_run1 & Albanian & -9.958 \\
154M+LT\_128-shuff-dyck\_run3 & Czech & -9.919 \\
308M+GT\_128-dyck\_run1 & Finnish & -9.912 \\
308M+GT\_64-shuff-dyck\_run1 & Finnish & -9.912 \\
308M+LT\_64-shuff-dyck\_run1 & Dutch & -9.908 \\
481M+GT\_64-dyck\_run3 & Dutch & -9.906 \\
481M+LT\_64-shuff-dyck\_run3 & English & -9.897 \\
154M+GT\_128-dyck\_run1 & Danish & -8.982 \\
481M+GT\_128-dyck\_run1 & Danish & -8.982 \\
481M+GT\_128-shuff-dyck\_run2 & Danish & -8.982 \\
308M+GT\_128-dyck\_run2 & Danish & -8.982 \\
481M+LT\_64-shuff-dyck\_run1 & Finnish & -8.925 \\
481M+LT\_64-dyck\_run1 & Danish & -8.915 \\
308M+GT\_128-shuff-dyck\_run3 & Finnish & -7.927 \\
308M+LT\_64-shuff-dyck\_run2 & Dutch & -7.909 \\
481M+GT\_64-dyck\_run2 & Dutch & -7.902 \\
481M+LT\_64-dyck\_run3 & English & -7.900 \\
481M+GT\_64-shuff-dyck\_run3 & Danish & -6.990 \\
481M+GT\_64-dyck\_run3 & Danish & -6.990 \\
154M+GT\_128-shuff-dyck\_run2 & Danish & -6.987 \\
308M+GT\_128-dyck\_run2 & Czech & -6.979 \\
481M+LT\_128-shuff-dyck\_run1 & Finnish & -6.927 \\
481M+LT\_128-dyck\_run3 & Dutch & -6.915 \\
481M+GT\_128-shuff-dyck\_run3 & Czech & -5.983 \\
481M+GT\_64-shuff-dyck\_run3 & Albanian & -5.971 \\
154M+LT\_128-shuff-dyck\_run1 & Albanian & -5.961 \\
481M+GT\_64-shuff-dyck\_run2 & Dutch & -5.094 \\
308M+GT\_128-dyck\_run3 & Finnish & -5.090 \\
481M+LT\_128-dyck\_run1 & Dutch & -5.089 \\
481M+LT\_64-shuff-dyck\_run2 & Dutch & -5.088 \\
308M+LT\_128-shuff-dyck\_run3 & Danish & -5.082 \\
154M+GT\_64-shuff-dyck\_run2 & English & -5.024 \\
308M+GT\_128-dyck\_run2 & Albanian & -4.974 \\
308M+LT\_128-dyck\_run1 & Dutch & -4.914 \\
481M+LT\_64-dyck\_run1 & Dutch & -4.914 \\
308M+GT\_64-dyck\_run3 & Dutch & -4.901 \\
481M+LT\_128-shuff-dyck\_run3 & Finnish & -4.087 \\
481M+GT\_64-dyck\_run2 & Danish & -4.009 \\
154M+GT\_128-shuff-dyck\_run1 & Danish & -3.980 \\
481M+LT\_64-shuff-dyck\_run2 & Danish & -3.927 \\
481M+LT\_128-dyck\_run2 & Danish & -3.927 \\
481M+LT\_64-shuff-dyck\_run3 & Dutch & -3.918 \\
308M+GT\_128-shuff-dyck\_run1 & Finnish & -3.910 \\
154M+GT\_64-dyck\_run1 & Dutch & -3.908 \\
481M+GT\_64-dyck\_run1 & Dutch & -3.908 \\
308M+LT\_64-shuff-dyck\_run1 & English & -3.885 \\
481M+GT\_64-dyck\_run1 & Finnish & -3.093 \\
308M+LT\_128-dyck\_run2 & Dutch & -3.085 \\
481M+LT\_128-dyck\_run3 & Danish & -3.081 \\
481M+GT\_64-dyck\_run1 & Albanian & -3.026 \\
481M+GT\_64-shuff-dyck\_run1 & Albanian & -3.026 \\
154M+GT\_128-dyck\_run1 & Albanian & -2.975 \\
154M+GT\_128-dyck\_run2 & Albanian & -2.974 \\
481M+LT\_128-shuff-dyck\_run3 & Czech & -2.920 \\
154M+GT\_64-shuff-dyck\_run2 & Dutch & -2.096 \\
154M+LT\_128-shuff-dyck\_run2 & Finnish & -2.072 \\
481M+LT\_128-dyck\_run2 & Finnish & -2.072 \\
481M+LT\_128-shuff-dyck\_run1 & Albanian & -2.040 \\
154M+GT\_128-shuff-dyck\_run3 & Albanian & -1.969 \\
308M+LT\_128-shuff-dyck\_run2 & Danish & -1.927 \\
481M+LT\_128-dyck\_run1 & Czech & -1.919 \\
481M+GT\_128-shuff-dyck\_run3 & Dutch & -1.917 \\
481M+GT\_64-dyck\_run2 & Finnish & -1.916 \\
308M+GT\_128-dyck\_run2 & Finnish & -1.916 \\
481M+LT\_64-shuff-dyck\_run3 & Danish & -1.913 \\
481M+LT\_64-shuff-dyck\_run1 & Dutch & -1.088 \\
308M+GT\_128-shuff-dyck\_run3 & English & -1.021 \\
154M+LT\_128-shuff-dyck\_run1 & Danish & -0.918 \\
481M+LT\_64-dyck\_run2 & Dutch & -0.906 \\
154M+LT\_64-dyck\_run3 & Dutch & -0.905 \\
481M+GT\_64-shuff-dyck\_run3 & Finnish & -0.073 \\
308M+GT\_128-shuff-dyck\_run1 & Albanian & -0.027 \\
154M+GT\_64-dyck\_run1 & Albanian & -0.027 \\
308M+GT\_64-shuff-dyck\_run3 & English & -0.026 \\
154M+GT\_64-dyck\_run1 & English & 0.016 \\
481M+GT\_128-shuff-dyck\_run1 & Albanian & 0.027 \\
308M+GT\_64-dyck\_run2 & Dutch & 0.104 \\
154M+LT\_128-dyck\_run3 & Czech & 0.910 \\
154M+GT\_64-dyck\_run1 & Czech & 1.021 \\
481M+LT\_128-shuff-dyck\_run1 & Dutch & 1.088 \\
481M+GT\_128-dyck\_run1 & Dutch & 1.094 \\
481M+GT\_128-shuff-dyck\_run2 & Dutch & 1.099 \\
308M+LT\_128-shuff-dyck\_run1 & English & 1.108 \\
308M+LT\_64-dyck\_run1 & English & 1.888 \\
481M+LT\_128-dyck\_run3 & English & 1.893 \\
481M+LT\_128-shuff-dyck\_run2 & English & 1.908 \\
481M+LT\_128-dyck\_run2 & Dutch & 1.915 \\
308M+GT\_64-shuff-dyck\_run3 & Dutch & 1.917 \\
154M+LT\_128-shuff-dyck\_run3 & Albanian & 1.955 \\
308M+LT\_128-shuff-dyck\_run1 & Albanian & 2.040 \\
154M+LT\_64-shuff-dyck\_run1 & Czech & 2.080 \\
481M+LT\_128-dyck\_run2 & English & 2.101 \\
308M+LT\_128-shuff-dyck\_run2 & English & 2.906 \\
481M+LT\_128-shuff-dyck\_run2 & Danish & 2.919 \\
308M+GT\_64-dyck\_run3 & Finnish & 2.925 \\
481M+LT\_128-dyck\_run2 & Czech & 2.935 \\
481M+LT\_128-dyck\_run1 & Albanian & 2.962 \\
154M+LT\_64-shuff-dyck\_run1 & Albanian & 2.962 \\
481M+GT\_128-dyck\_run3 & Danish & 2.977 \\
481M+GT\_64-shuff-dyck\_run1 & Czech & 2.979 \\
154M+GT\_64-dyck\_run2 & English & 3.020 \\
154M+GT\_128-shuff-dyck\_run2 & English & 3.020 \\
481M+LT\_128-shuff-dyck\_run2 & Dutch & 3.085 \\
308M+GT\_64-shuff-dyck\_run2 & Finnish & 3.904 \\
308M+GT\_128-dyck\_run1 & Dutch & 3.908 \\
154M+LT\_64-shuff-dyck\_run3 & Albanian & 3.956 \\
308M+GT\_64-shuff-dyck\_run1 & Albanian & 3.975 \\
154M+GT\_128-dyck\_run3 & English & 3.981 \\
308M+GT\_128-dyck\_run1 & English & 3.983 \\
481M+GT\_64-shuff-dyck\_run2 & Danish & 3.988 \\
154M+LT\_64-dyck\_run2 & Czech & 4.078 \\
308M+GT\_128-shuff-dyck\_run2 & Dutch & 4.902 \\
154M+GT\_64-dyck\_run3 & Finnish & 4.921 \\
154M+GT\_64-shuff-dyck\_run2 & Czech & 4.975 \\
154M+LT\_64-shuff-dyck\_run3 & Czech & 5.072 \\
154M+LT\_64-dyck\_run3 & Czech & 5.072 \\
308M+LT\_128-dyck\_run2 & English & 5.901 \\
154M+GT\_64-shuff-dyck\_run3 & English & 5.980 \\
154M+GT\_64-dyck\_run1 & Danish & 5.982 \\
308M+GT\_64-shuff-dyck\_run2 & Dutch & 6.907 \\
308M+LT\_64-shuff-dyck\_run2 & Albanian & 6.946 \\
308M+LT\_128-dyck\_run1 & Albanian & 6.961 \\
154M+GT\_128-dyck\_run2 & English & 6.978 \\
154M+GT\_64-shuff-dyck\_run1 & English & 6.983 \\
154M+GT\_128-dyck\_run2 & Danish & 6.987 \\
308M+LT\_128-dyck\_run1 & English & 7.891 \\
481M+LT\_64-dyck\_run1 & English & 7.891 \\
481M+LT\_64-shuff-dyck\_run1 & English & 7.891 \\
154M+LT\_64-shuff-dyck\_run1 & Dutch & 7.910 \\
481M+LT\_64-shuff-dyck\_run3 & Czech & 7.912 \\
481M+LT\_128-shuff-dyck\_run3 & Danish & 7.914 \\
154M+LT\_128-shuff-dyck\_run2 & Czech & 7.935 \\
481M+LT\_128-shuff-dyck\_run2 & Finnish & 7.940 \\
154M+GT\_128-dyck\_run2 & Czech & 7.975 \\
154M+GT\_64-dyck\_run3 & English & 7.980 \\
481M+LT\_128-shuff-dyck\_run3 & English & 8.899 \\
308M+GT\_128-shuff-dyck\_run1 & Dutch & 8.903 \\
154M+GT\_128-shuff-dyck\_run3 & Dutch & 8.910 \\
481M+LT\_64-shuff-dyck\_run1 & Czech & 8.920 \\
481M+LT\_64-dyck\_run3 & Danish & 8.922 \\
481M+LT\_64-dyck\_run3 & Albanian & 8.962 \\
154M+GT\_64-shuff-dyck\_run1 & Albanian & 8.976 \\
308M+GT\_128-dyck\_run3 & Albanian & 8.977 \\
308M+GT\_128-dyck\_run2 & English & 8.977 \\
154M+GT\_128-shuff-dyck\_run3 & English & 8.979 \\
308M+GT\_64-shuff-dyck\_run1 & English & 8.983 \\
308M+LT\_128-shuff-dyck\_run3 & English & 9.897 \\
308M+GT\_128-dyck\_run3 & Dutch & 9.906 \\
154M+LT\_128-shuff-dyck\_run2 & English & 9.908 \\
154M+LT\_128-dyck\_run3 & Dutch & 9.910 \\
308M+GT\_128-shuff-dyck\_run1 & English & 9.982 \\
154M+GT\_64-shuff-dyck\_run3 & Danish & 9.986 \\
308M+GT\_64-shuff-dyck\_run3 & Danish & 9.993 \\
308M+LT\_64-shuff-dyck\_run3 & English & 10.896 \\
308M+LT\_64-shuff-dyck\_run2 & English & 10.908 \\
308M+LT\_128-shuff-dyck\_run1 & Dutch & 10.908 \\
154M+GT\_128-shuff-dyck\_run3 & Finnish & 10.918 \\
308M+LT\_64-shuff-dyck\_run2 & Czech & 10.933 \\
481M+LT\_64-shuff-dyck\_run2 & Czech & 10.933 \\
308M+LT\_128-shuff-dyck\_run2 & Albanian & 10.947 \\
154M+GT\_64-dyck\_run2 & Albanian & 10.981 \\
481M+LT\_128-dyck\_run1 & English & 11.889 \\
308M+GT\_128-shuff-dyck\_run3 & Dutch & 11.910 \\
481M+GT\_128-dyck\_run2 & Dutch & 11.912 \\
308M+LT\_128-shuff-dyck\_run1 & Finnish & 11.926 \\
154M+LT\_64-shuff-dyck\_run1 & Finnish & 11.926 \\
154M+LT\_64-dyck\_run1 & Albanian & 11.957 \\
154M+GT\_128-dyck\_run3 & Danish & 11.986 \\
308M+LT\_64-dyck\_run3 & English & 12.893 \\
308M+LT\_128-dyck\_run3 & English & 12.893 \\
308M+LT\_64-dyck\_run1 & Dutch & 12.904 \\
481M+GT\_128-shuff-dyck\_run1 & Dutch & 12.905 \\
481M+GT\_64-shuff-dyck\_run1 & Dutch & 12.905 \\
154M+GT\_64-dyck\_run3 & Dutch & 12.914 \\
481M+LT\_64-dyck\_run3 & Czech & 12.920 \\
308M+LT\_64-dyck\_run3 & Danish & 12.924 \\
154M+LT\_128-dyck\_run2 & Albanian & 12.945 \\
308M+LT\_128-dyck\_run2 & Albanian & 12.947 \\
154M+LT\_64-dyck\_run3 & Albanian & 12.955 \\
308M+GT\_128-shuff-dyck\_run2 & English & 12.974 \\
481M+GT\_128-dyck\_run2 & Danish & 12.980 \\
154M+GT\_64-shuff-dyck\_run2 & Danish & 12.985 \\
154M+GT\_128-shuff-dyck\_run3 & Danish & 12.986 \\
308M+LT\_64-dyck\_run1 & Danish & 13.911 \\
154M+LT\_64-shuff-dyck\_run2 & English & 13.912 \\
308M+LT\_64-dyck\_run2 & English & 13.919 \\
308M+GT\_64-shuff-dyck\_run3 & Finnish & 13.922 \\
481M+LT\_128-shuff-dyck\_run1 & Czech & 13.926 \\
481M+LT\_64-shuff-dyck\_run2 & Finnish & 13.939 \\
308M+LT\_64-dyck\_run2 & Albanian & 13.948 \\
481M+GT\_64-dyck\_run3 & Albanian & 13.977 \\
308M+GT\_64-dyck\_run3 & English & 13.984 \\
308M+GT\_64-shuff-dyck\_run2 & English & 13.984 \\
154M+GT\_64-dyck\_run2 & Danish & 13.985 \\
308M+GT\_128-dyck\_run2 & Dutch & 14.897 \\
154M+LT\_64-dyck\_run2 & English & 14.910 \\
308M+LT\_64-shuff-dyck\_run3 & Dutch & 14.912 \\
481M+GT\_128-shuff-dyck\_run1 & Finnish & 14.913 \\
481M+LT\_64-dyck\_run2 & Danish & 14.917 \\
154M+LT\_128-shuff-dyck\_run1 & Finnish & 14.931 \\
154M+LT\_64-shuff-dyck\_run2 & Finnish & 14.933 \\
154M+LT\_128-shuff-dyck\_run1 & English & 15.891 \\
154M+GT\_128-dyck\_run1 & Dutch & 15.910 \\
481M+GT\_128-dyck\_run1 & Finnish & 15.915 \\
481M+GT\_64-shuff-dyck\_run1 & Finnish & 15.915 \\
308M+GT\_64-dyck\_run1 & Finnish & 15.915 \\
308M+LT\_128-shuff-dyck\_run1 & Czech & 15.926 \\
308M+LT\_64-dyck\_run1 & Finnish & 15.927 \\
481M+GT\_64-dyck\_run1 & English & 15.983 \\
481M+GT\_128-shuff-dyck\_run1 & English & 15.983 \\
308M+GT\_64-shuff-dyck\_run1 & Dutch & 16.911 \\
154M+GT\_64-shuff-dyck\_run1 & Dutch & 16.911 \\
308M+LT\_128-shuff-dyck\_run2 & Dutch & 16.915 \\
308M+LT\_64-dyck\_run2 & Dutch & 16.915 \\
154M+LT\_128-dyck\_run2 & Dutch & 16.917 \\
308M+LT\_128-dyck\_run1 & Finnish & 16.926 \\
481M+LT\_128-shuff-dyck\_run2 & Czech & 16.933 \\
481M+GT\_64-shuff-dyck\_run1 & English & 16.983 \\
308M+GT\_128-dyck\_run3 & English & 16.985 \\
154M+GT\_128-dyck\_run2 & Dutch & 17.904 \\
308M+LT\_128-dyck\_run3 & Dutch & 17.909 \\
154M+LT\_128-dyck\_run2 & English & 17.911 \\
154M+LT\_128-shuff-dyck\_run3 & Danish & 17.919 \\
308M+LT\_128-shuff-dyck\_run2 & Czech & 17.925 \\
154M+GT\_128-shuff-dyck\_run1 & Albanian & 17.976 \\
481M+GT\_128-dyck\_run1 & English & 17.983 \\
154M+LT\_64-dyck\_run1 & English & 18.884 \\
481M+LT\_128-shuff-dyck\_run1 & English & 18.884 \\
154M+LT\_64-dyck\_run1 & Dutch & 18.906 \\
308M+LT\_128-shuff-dyck\_run1 & Danish & 18.918 \\
308M+LT\_128-shuff-dyck\_run3 & Dutch & 18.918 \\
308M+LT\_64-dyck\_run3 & Dutch & 18.918 \\
154M+LT\_128-dyck\_run1 & Czech & 18.921 \\
481M+LT\_64-dyck\_run1 & Czech & 18.921 \\
154M+LT\_64-shuff-dyck\_run2 & Czech & 18.928 \\
154M+LT\_128-dyck\_run1 & Albanian & 18.954 \\
154M+GT\_64-dyck\_run2 & Czech & 18.976 \\
154M+GT\_64-shuff-dyck\_run2 & Albanian & 18.977 \\
154M+LT\_64-shuff-dyck\_run1 & English & 19.882 \\
154M+LT\_128-dyck\_run1 & English & 19.882 \\
154M+LT\_128-shuff-dyck\_run3 & English & 19.895 \\
154M+LT\_64-dyck\_run3 & English & 19.895 \\
154M+GT\_64-shuff-dyck\_run3 & Dutch & 19.910 \\
308M+LT\_64-shuff-dyck\_run3 & Czech & 19.912 \\
308M+LT\_64-shuff-dyck\_run1 & Czech & 19.922 \\
308M+LT\_128-dyck\_run2 & Danish & 19.924 \\
154M+LT\_128-dyck\_run2 & Czech & 19.926 \\
154M+GT\_128-shuff-dyck\_run1 & English & 19.982 \\
154M+GT\_128-dyck\_run1 & English & 19.982 \\
154M+GT\_64-dyck\_run3 & Danish & 19.985 \\
154M+LT\_128-dyck\_run1 & Dutch & 20.907 \\
154M+LT\_64-dyck\_run2 & Dutch & 20.916 \\
308M+LT\_128-dyck\_run1 & Danish & 20.922 \\
308M+LT\_64-dyck\_run2 & Finnish & 20.929 \\
154M+LT\_64-shuff-dyck\_run3 & Finnish & 20.930 \\
481M+GT\_128-dyck\_run3 & English & 20.982 \\
154M+LT\_64-shuff-dyck\_run3 & English & 21.893 \\
154M+GT\_128-dyck\_run3 & Dutch & 21.910 \\
308M+LT\_128-dyck\_run3 & Danish & 21.919 \\
154M+LT\_64-dyck\_run1 & Czech & 21.922 \\
481M+LT\_64-dyck\_run2 & Czech & 21.924 \\
154M+LT\_128-dyck\_run1 & Finnish & 21.929 \\
154M+LT\_64-dyck\_run1 & Finnish & 21.929 \\
154M+GT\_128-shuff-dyck\_run2 & Albanian & 21.979 \\
481M+GT\_128-shuff-dyck\_run1 & Czech & 21.979 \\
154M+LT\_128-shuff-dyck\_run2 & Danish & 22.926 \\
308M+LT\_64-shuff-dyck\_run2 & Finnish & 22.928 \\
154M+LT\_128-shuff-dyck\_run3 & Finnish & 22.930 \\
154M+LT\_64-dyck\_run2 & Finnish & 22.932 \\
154M+LT\_128-dyck\_run2 & Finnish & 22.932 \\
154M+LT\_128-dyck\_run3 & English & 23.892 \\
308M+LT\_64-dyck\_run1 & Czech & 23.922 \\
308M+LT\_64-shuff-dyck\_run2 & Danish & 23.924 \\
308M+LT\_64-dyck\_run2 & Czech & 23.924 \\
308M+LT\_64-shuff-dyck\_run3 & Finnish & 23.936 \\
308M+LT\_64-dyck\_run3 & Finnish & 23.936 \\
308M+LT\_128-shuff-dyck\_run2 & Finnish & 23.936 \\
308M+LT\_128-shuff-dyck\_run3 & Finnish & 23.936 \\
154M+LT\_128-dyck\_run3 & Albanian & 23.959 \\
308M+LT\_128-dyck\_run2 & Czech & 24.916 \\
154M+LT\_128-shuff-dyck\_run1 & Czech & 25.920 \\
308M+LT\_64-shuff-dyck\_run3 & Danish & 25.920 \\
308M+LT\_64-dyck\_run2 & Danish & 25.923 \\
308M+LT\_64-shuff-dyck\_run1 & Danish & 25.924 \\
308M+LT\_64-shuff-dyck\_run1 & Finnish & 25.927 \\
308M+LT\_128-shuff-dyck\_run3 & Czech & 25.927 \\
154M+LT\_64-dyck\_run3 & Finnish & 25.931 \\
308M+LT\_128-dyck\_run2 & Finnish & 25.935 \\
308M+LT\_128-dyck\_run3 & Finnish & 26.927 \\
308M+LT\_128-dyck\_run1 & Czech & 27.921 \\
154M+LT\_128-dyck\_run3 & Danish & 27.923 \\
308M+LT\_128-dyck\_run3 & Czech & 28.920 \\
308M+LT\_64-dyck\_run3 & Czech & 28.920 \\
154M+LT\_64-shuff-dyck\_run3 & Danish & 28.923 \\
154M+LT\_64-shuff-dyck\_run1 & Danish & 28.924 \\
154M+LT\_64-dyck\_run1 & Danish & 28.924 \\
154M+LT\_64-shuff-dyck\_run2 & Danish & 29.917 \\
154M+LT\_128-dyck\_run2 & Danish & 29.917 \\
154M+LT\_64-dyck\_run2 & Danish & 29.917 \\
154M+LT\_128-dyck\_run3 & Finnish & 29.929 \\
154M+LT\_128-dyck\_run1 & Danish & 30.923 \\
154M+LT\_64-dyck\_run3 & Danish & 31.920
\cr\hline
\end{supertabular}
\end{small}

\clearpage
\section{Description of Linguistic Features}
\label{app:C}
We estimate 11 linguistic metrics from 1M token samples per language from training and test data. One additional exception feature is  
\circle{10} \textbf{Morphological richness} which is estimated on the Bible dataset \cite{christodouloupoulos2015massively}, covering languages in a multi-parallel manner.

Below is a detailed list of linguistic features we analyze. Most features are considered in two variants: (1) punctuation is taken into account when processing the data (\texttt{*\_with\_punct}); and (2) punctuation is ignored (\texttt{*\_without\_punct}).  The head of a punctuation token is typically the syntactic head of the clause or the phrase it punctuates, which can make the properties replicate measures of sentence length rather than providing structural insights. 

We use the multi-lingual \texttt{gemma-3-12B-pt} model
\citep{gemmateam2025gemma3technicalreport} to measure \circle{11} \textbf{Average perplexity} and  \circle{12} \textbf{Variance of the perplexity} to quantify the predictability and repetitiveness of data (i.e., higher values should indicate more diverse data).


\begin{enumerate}[topsep=0.5ex,itemsep=0ex,partopsep=0ex]
    \item[\circle{1}] \textbf{Percentage of crossing dependencies:}  ratio of crossing dependent pairs to all dependent pairs. \\(\texttt{\%crossing\_dep\_with\_punct; \\ \%crossing\_dep\_without\_punct})  
    \item[\circle{2}] \textbf{Mean dependency distance:} average distance between the head and the child in the dependent pair. \\(\texttt{mean\_dep\_dist\_with\_punct; \\ mean\_dep\_dist\_without\_punct})
    \item[\circle{3}] \textbf{Maximum dependency distance:} maximum distance between the head and the child in the dependent pair. \\(\texttt{max\_dep\_dist\_with\_punct; \\ max\_dep\_dist\_without\_punct})
    \item[\circle{4}] \textbf{Mean dependency tree depth:} average depth of the dependency tree. \\(\texttt{mean\_tree\_depth\_with\_punct; \\ mean\_tree\_depth\_without\_punct})
    \item[\circle{5}] \textbf{Maximum dependency tree depth:} maximum depth of the dependency tree. \\(\texttt{max\_tree\_depth\_with\_punct; \\ max\_tree\_depth\_without\_punct})
    \item[\circle{6}] \textbf{Vocabulary size:} the number of unique tokens. \\(\texttt{vocab\_size\_with\_punct; \\ vocab\_size\_without\_punct})
    \item[\circle{7}] \textbf{Average sentence length:} average number of tokens in the sentence. \\(\texttt{avg\_sentence\_length\_with\_punct; \\ avg\_sentence\_length\_without\_punct})
    \item[\circle{8}] \textbf{Average number of children per head:} average number of children per head across all dependency pairs. \\(\texttt{mean\_children\_with\_punct; \\ mean\_children\_without\_punct})
    \item[\circle{9}] \textbf{Maximum number of children per head:} maximum number of children per head across all dependency pairs. \\(\texttt{max\_children\_with\_punct; \\ max\_children\_without\_punct})
    \item[\circle{10}] \textbf{Morphological richness:} the number of distinct word forms seen in a parallel corpus relative to English \cite{10.1093/oxfordhb/9780199591428.013.13}. \\(\texttt{morph\_richness})
    \item[\circle{11}] \textbf{Average perplexity:} the average of perplexities measured on batches of 2,048 tokens. \\ (\texttt{avg\_perplexity})
    \item[\circle{12}] \textbf{Variance of the perplexity:} the variance of perplexities measured on batches of 2,048 tokens. \\(\texttt{var\_perplexity})
\end{enumerate}


\section{Details of Language Specific Analysis}
\label{app:D}

\begin{figure*}[h!]
    \centering
    \includegraphics[width=0.9\linewidth]{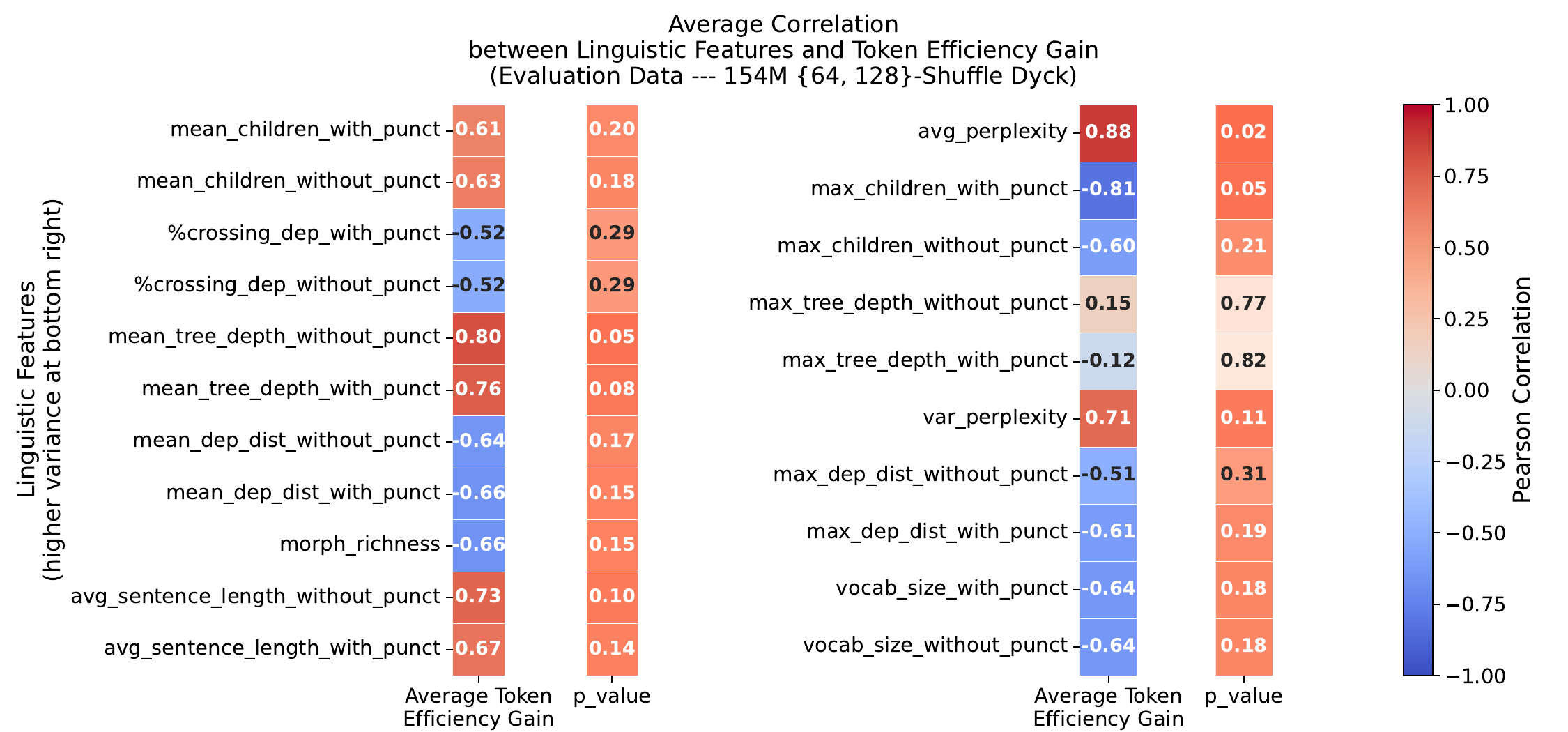}
    \caption{Average correlation between linguistic features of evaluation data and token efficiency gain of 154M models pretrained on \{64,128\}-Shuffle Dyck. \texttt{p-values} have been corrected with the false discovery rate method.}
    \label{fig:average_correlation_eval_shuff}
\end{figure*}

\begin{figure*}[h!]
    \centering
    \includegraphics[width=0.9\linewidth]{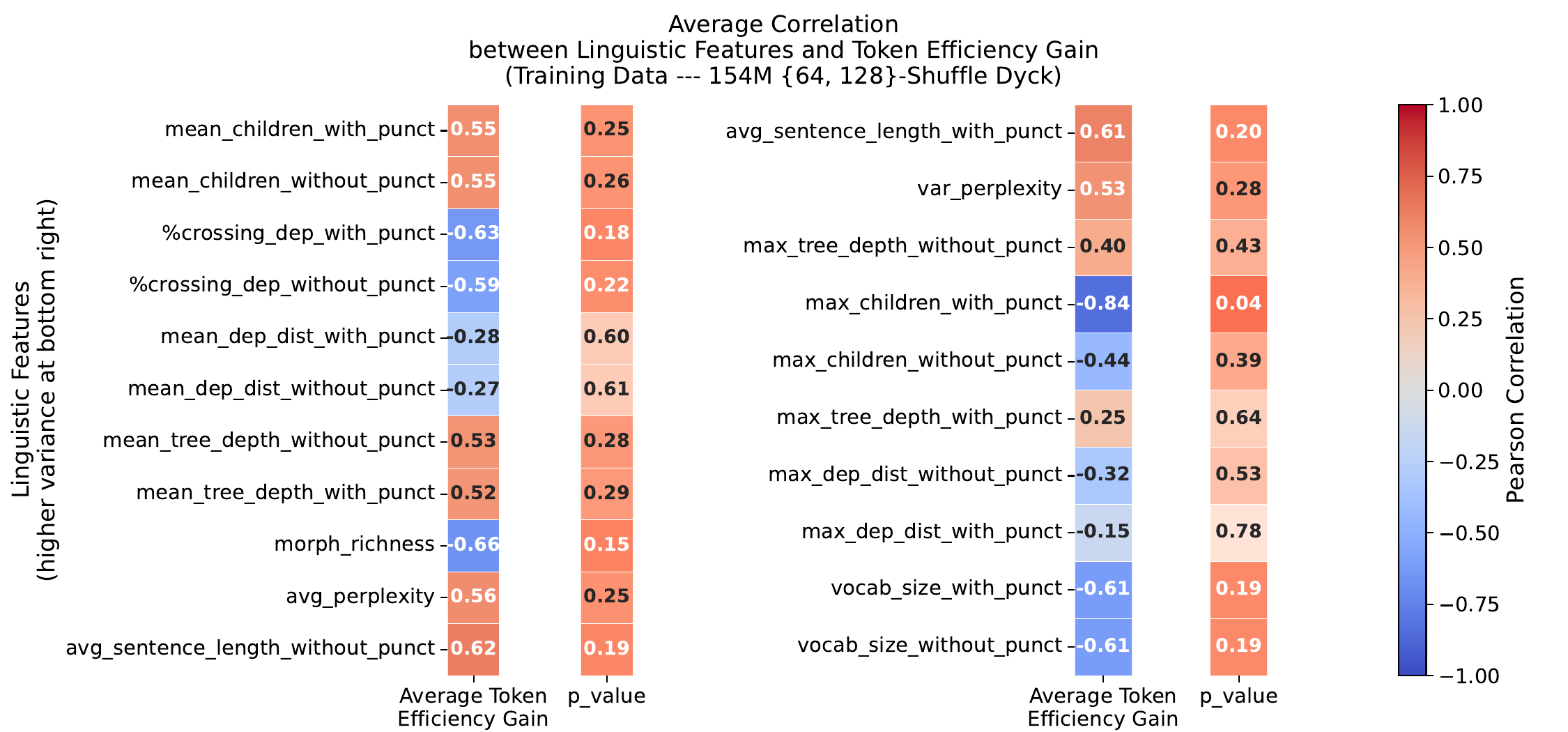}
    \caption{Average correlation between linguistic features of training data and token efficiency gain of 154M models on \{64,128\}-Shuffle Dyck. \texttt{p-values} have been corrected with the false discovery rate method.}
    \label{fig:average_correlation_train_shuff}
\end{figure*}

\cref{fig:average_correlation_eval_shuff,fig:average_correlation_train_shuff} provide the average correlations between efficiency gain across models pretrained on \{64,128\}-Shuffle Dyck and linguistic features (see \autoref{app:C}) estimated on evaluation and training data, respectively. 

\begin{figure*}[h!]
    \centering
    \includegraphics[width=0.9\linewidth]{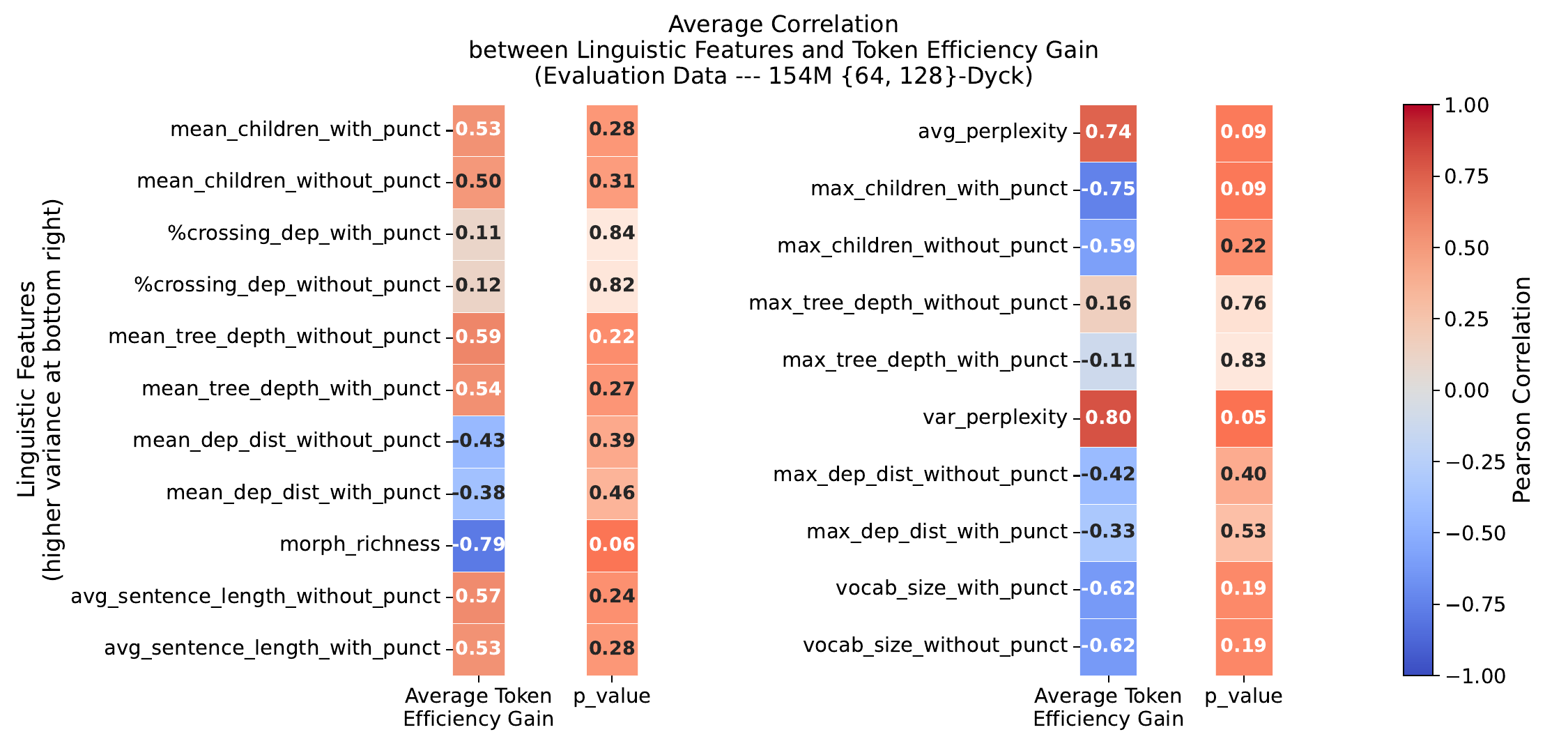}
    \caption{Average correlation between linguistic features of evaluation data and token efficiency gain of 154M models pretrained on \{64,128\}-Dyck. \texttt{p-values} have been corrected with the false discovery rate method.}
    \label{fig:average_correlation_eval_dyck}
\end{figure*}

\begin{figure*}[h!]
    \centering
    \includegraphics[width=0.9\linewidth]{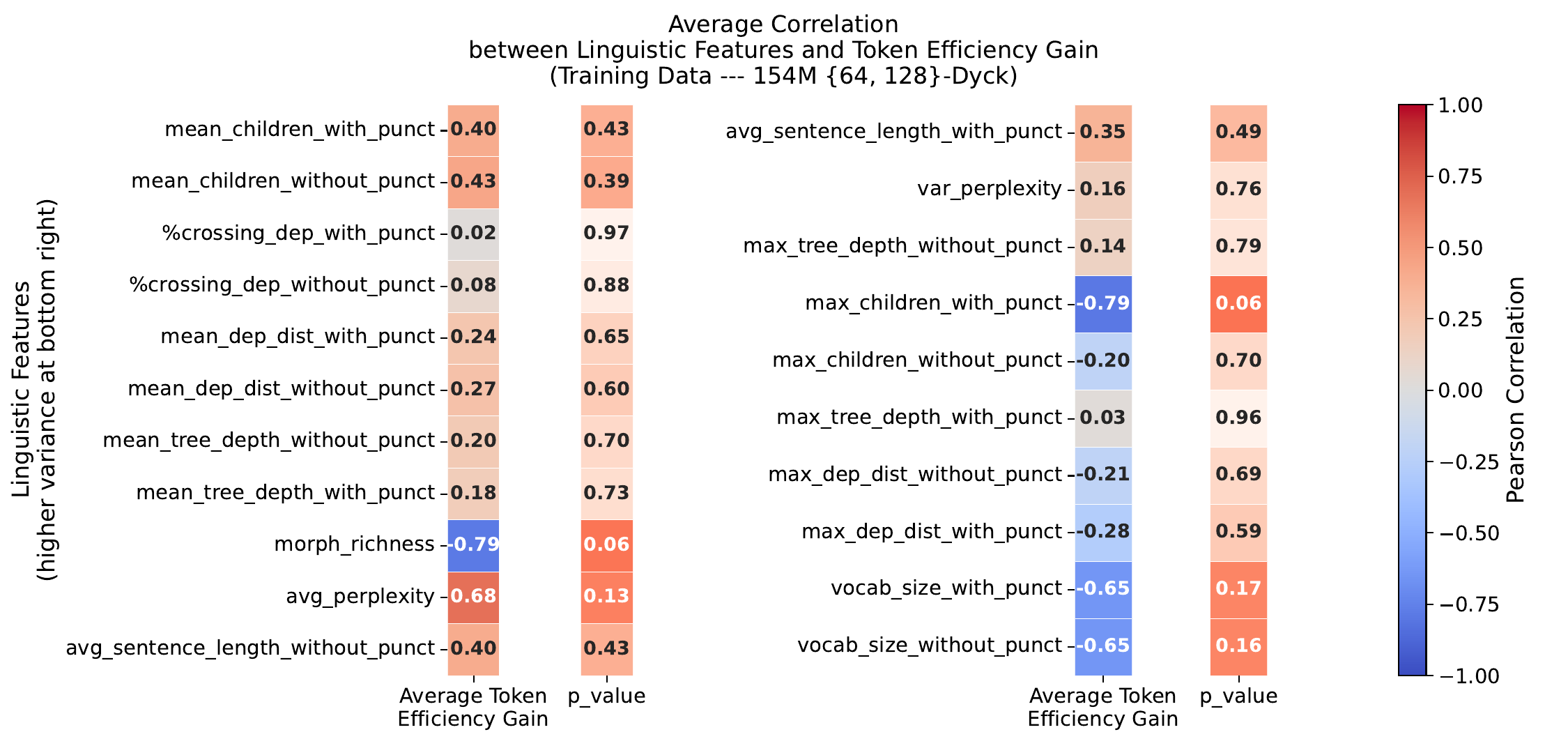}
    \caption{Average correlation between linguistic features of training data and token efficiency gain of 154M models on \{64,128\}-Dyck. \texttt{p-values} have been corrected with the false discovery rate method.}
    \label{fig:average_correlation_train_dyck}
\end{figure*}

\cref{fig:average_correlation_eval_dyck,fig:average_correlation_train_dyck} provide the average correlations between efficiency gain across models pretrained on \{64,128\}-Dyck and linguistic features (see \autoref{app:C}) estimated on evaluation and training data, respectively. 

\begin{figure*}[h!]
    \centering
    \includegraphics[width=0.9\linewidth]{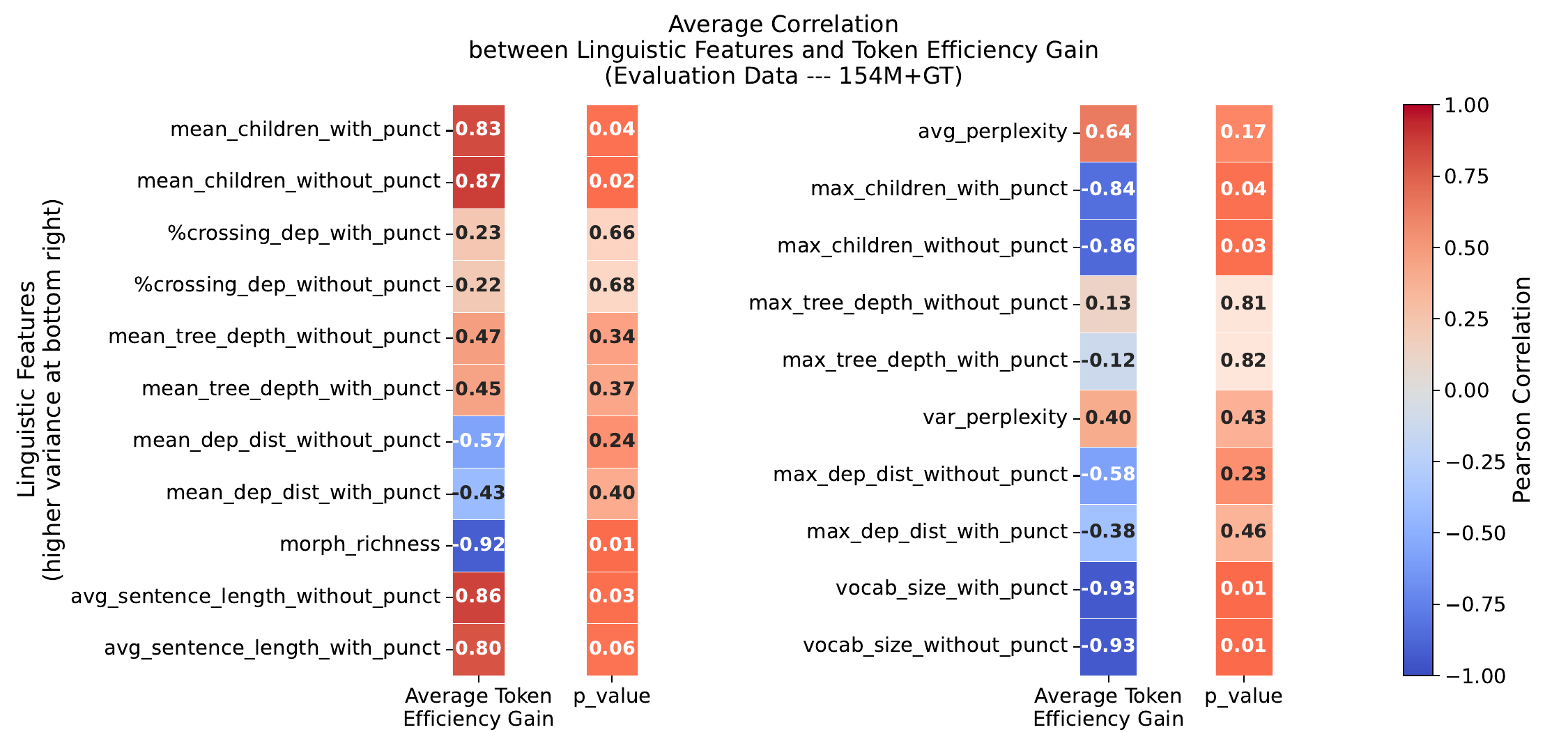}
    \caption{Average correlation between linguistic features of evaluation data and token efficiency gain of 154M models with the Gemma tokenizer. \texttt{p-values} have been corrected with the false discovery rate method.}
    \label{fig:average_correlation_GT_eval}
\end{figure*}

\begin{figure*}[h!]
    \centering
    \includegraphics[width=0.9\linewidth]{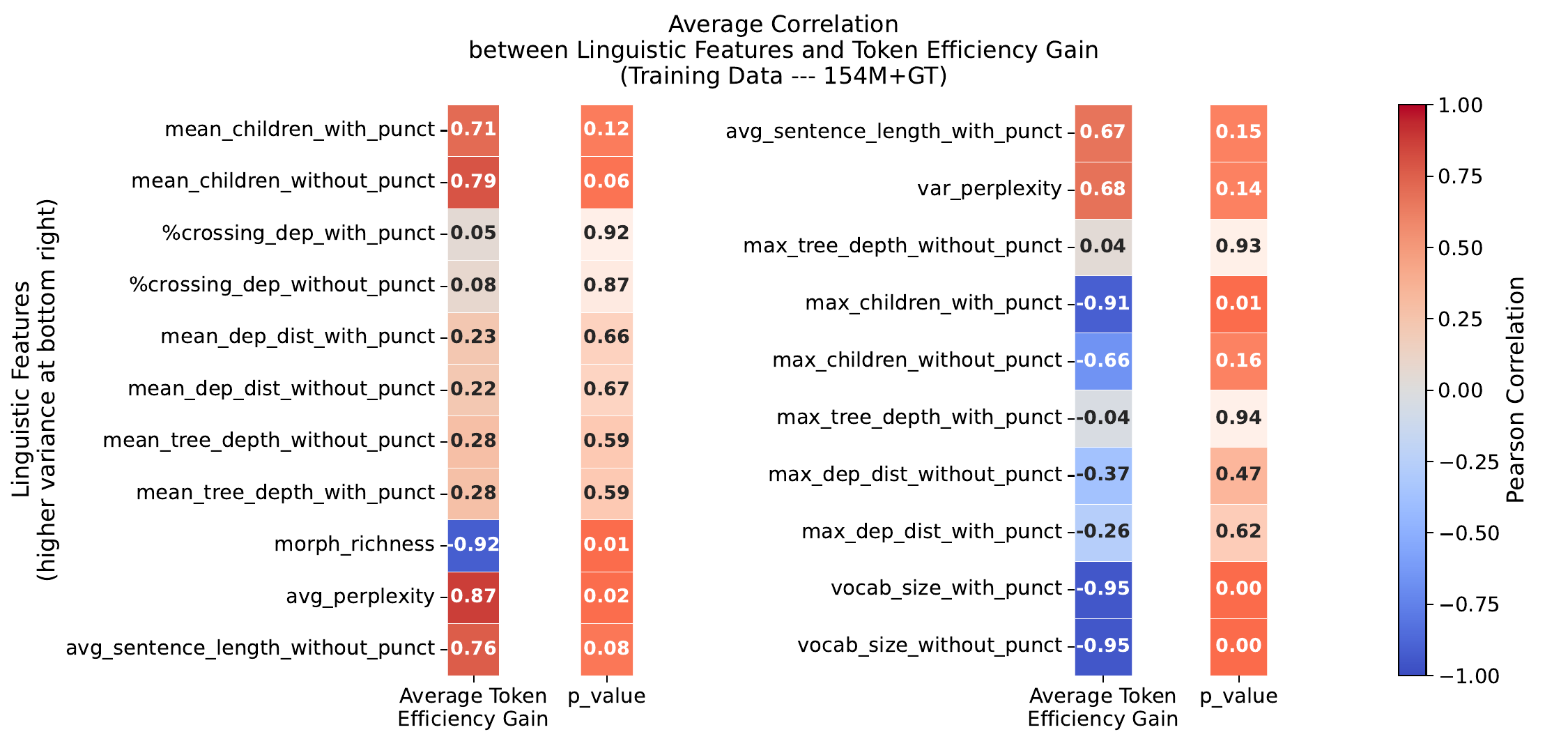}
    \caption{Average correlation between linguistic features of training data and token efficiency gain of 154M models with the Gemma tokenizer. \texttt{p-values} have been corrected with the false discovery rate method.}
    \label{fig:average_correlation_GT_train}
\end{figure*}

\cref{fig:average_correlation_GT_eval,fig:average_correlation_GT_train} provide the average correlations between efficiency gain across models with the Gemma tokenizer and linguistic features (see \autoref{app:C}) estimated on evaluation and training data, respectively.

\begin{figure*}[h!]
    \centering
    \includegraphics[width=0.9\linewidth]{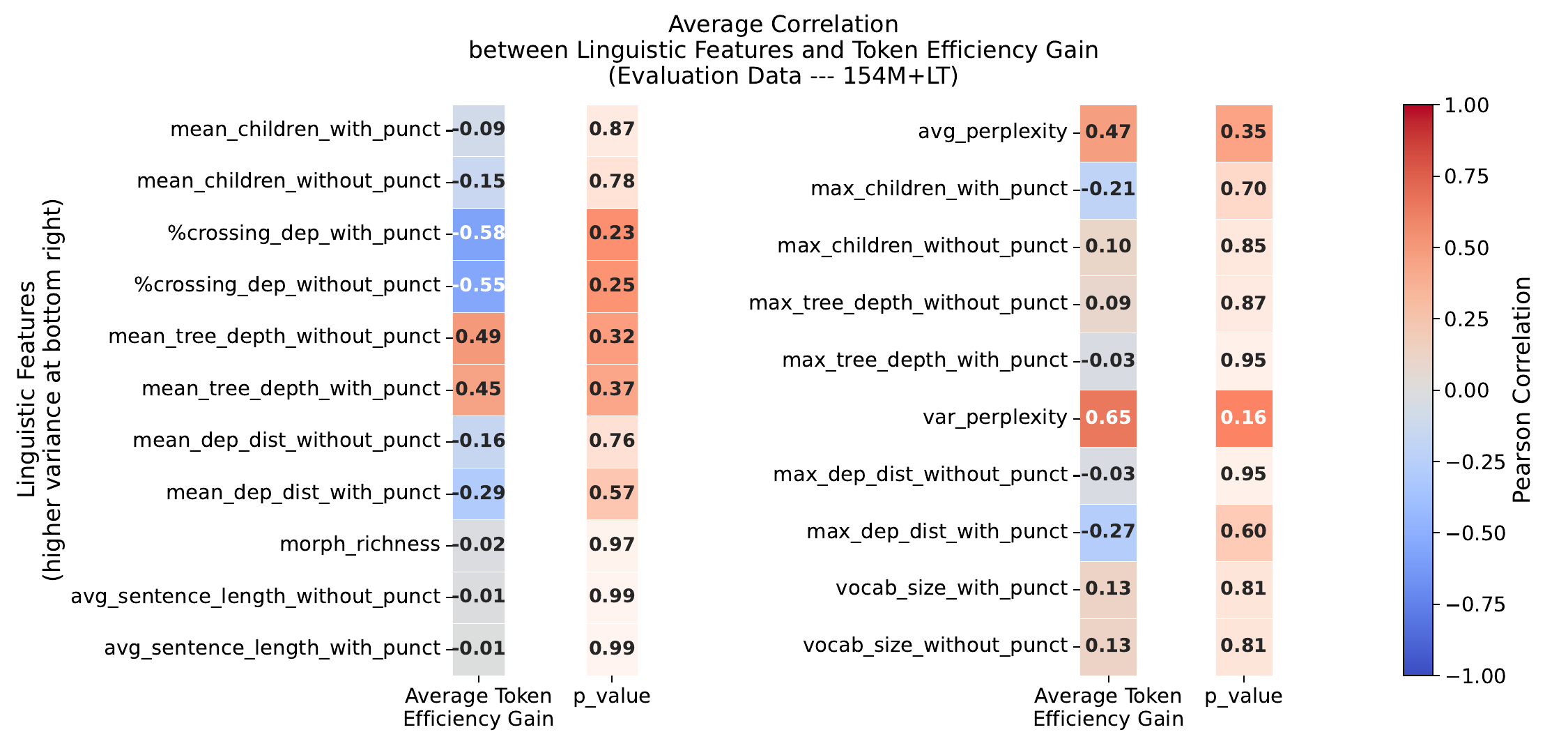}
    \caption{Average correlation between linguistic features of evaluation data and token efficiency gain of 154M models with the Llama tokenizer. \texttt{p\_values} have been corrected with the false discovery rate method.}
    \label{fig:average_correlation_LT_eval}
\end{figure*}
\begin{figure*}[h!]
    \centering
    \includegraphics[width=0.9\linewidth]{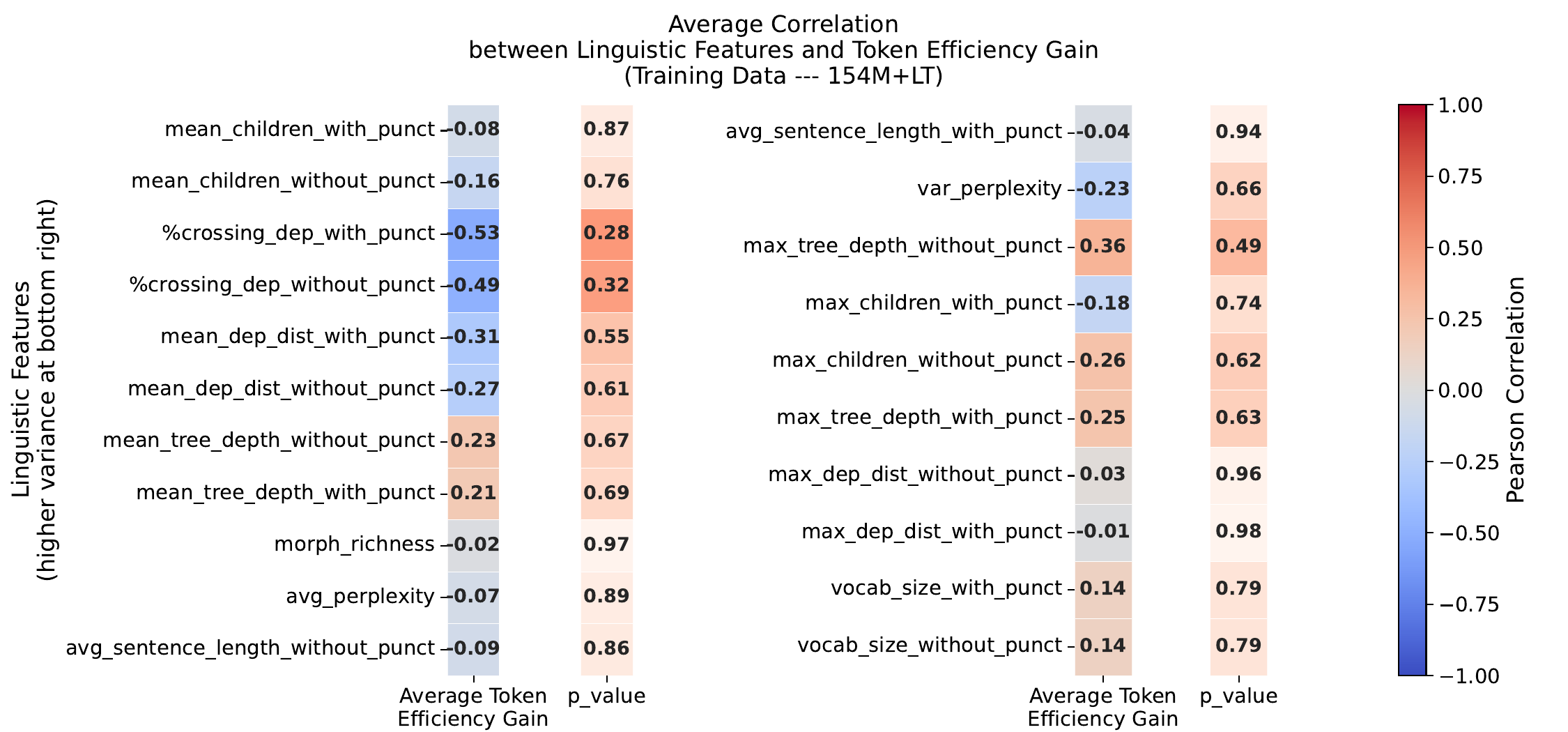}
    \caption{Average correlation between linguistic features of training data and token efficiency gain of 154M models with the Llama tokenizer. \texttt{p-values} have been corrected with the false discovery rate method.}
    \label{fig:average_correlation_LT_train}
\end{figure*}

\cref{fig:average_correlation_LT_eval,fig:average_correlation_LT_train} provide the average correlations between efficiency gain across models with the Llama tokenizer and linguistic features (see \autoref{app:C}) estimated on evaluation and training data, respectively. 

\begin{figure*}[h!]
    \centering
    \includegraphics[height=0.8\paperheight]{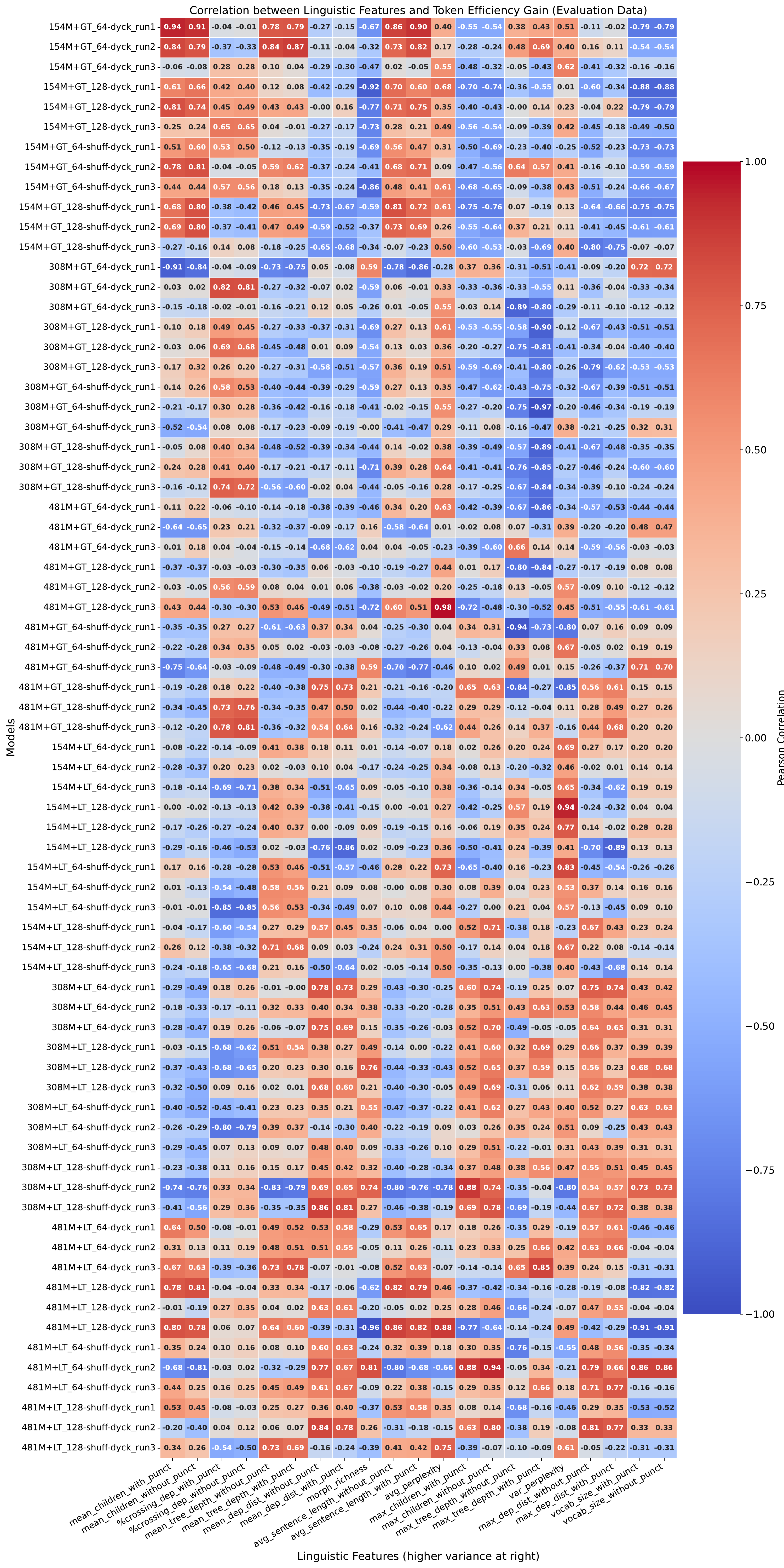}
    \caption{Correlation between linguistic features and token efficiency gain (evaluation data)}
    \label{fig:correlations_all_eval}
\end{figure*}

\begin{figure*}[h!]
    \centering
    \includegraphics[height=0.8\paperheight]{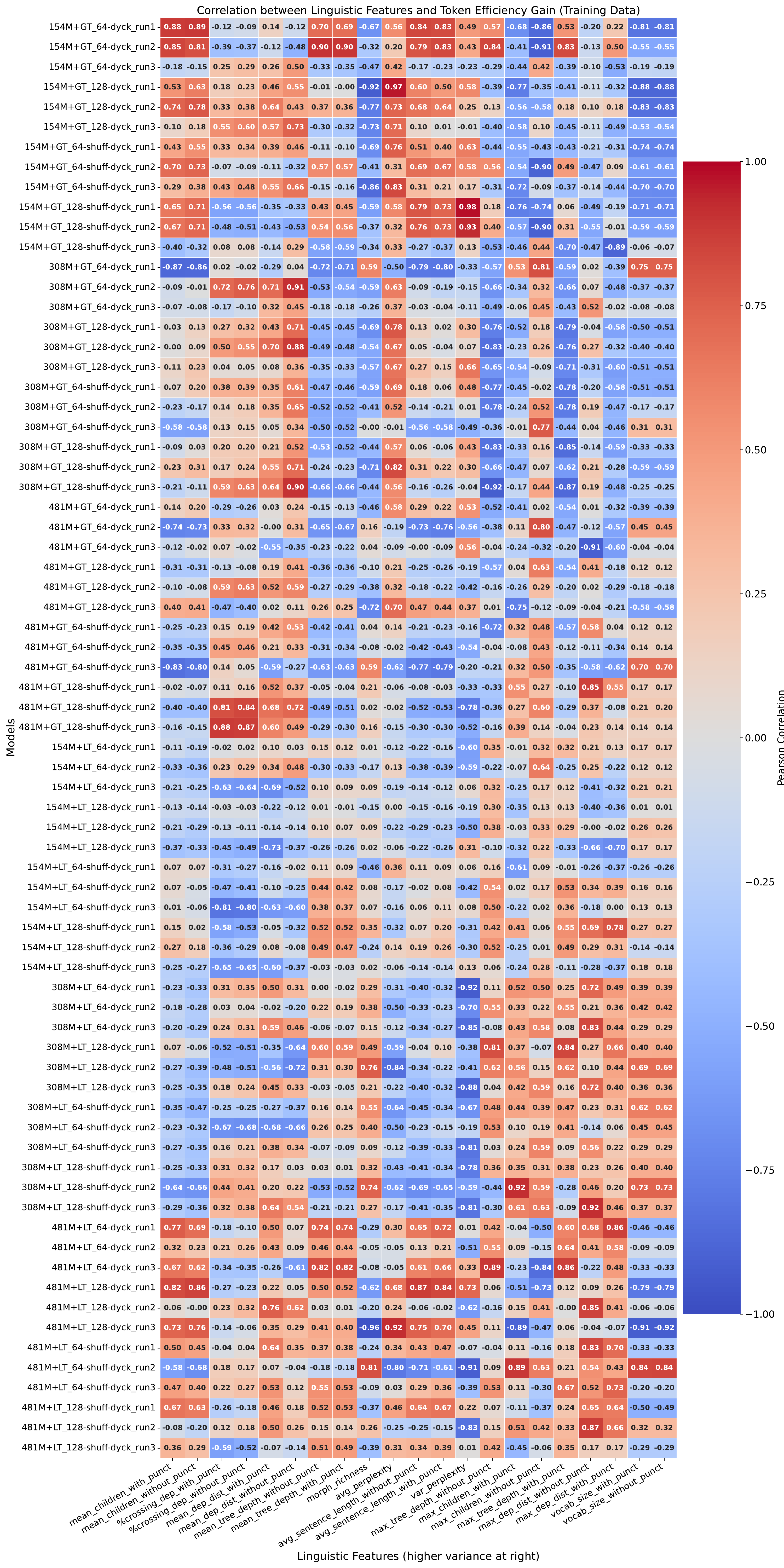}
    \caption{Correlation between linguistic features and token efficiency gain (training data)}
    \label{fig:correlations_all_train}
\end{figure*}

\cref{fig:correlations_all_eval,fig:correlations_all_train} provide the correlations between efficiency gain across experiments and linguistic features (see \autoref{app:C}) estimated on evaluation and training data, respectively.

Linguistic features are sorted from top to bottom and from left to right (i.e., higher variance at the bottom right side) for 
\cref{fig:average_correlation_eval_shuff,fig:average_correlation_train_shuff,fig:average_correlation_eval_dyck,fig:average_correlation_train_dyck,fig:average_correlation_GT_eval,fig:average_correlation_GT_train,fig:average_correlation_LT_eval,fig:average_correlation_LT_train},
 and from left to right (i.e., higher variance at the right side) for \cref{fig:correlations_all_eval,fig:correlations_all_train}. They are sorted based on their reliability which is quantified by the variance of the feature value across selected natural languages.

\clearpage





\end{document}